\documentclass[preprint,12pt]{elsarticle}

\usepackage{amssymb}
\usepackage{subcaption}
\usepackage{graphicx}
\usepackage{float}

\usepackage[]{tikz}
\usetikzlibrary{positioning}
\usetikzlibrary{shapes.multipart}
\usetikzlibrary{arrows.meta,fit,calc}

\usepackage{amsmath, amsbsy}
\usepackage{multirow, booktabs}

\journal{Alexandria Engineering Journal}

\begin{document}

\begin{frontmatter}



\title{Start Small: Training Controllable Game Level Generators without Training Data by Learning at Multiple Sizes}


\author{Yahia Zakaria, Magda Fayek, Mayada Hadhoud}

\affiliation{organization={Computer Engineering Department, Faculty of Engineering, Cairo University},%Department and Organization
            addressline={1 Gamaa Street}, 
            city={Giza},
            postcode={12613}, 
            state={Giza},
            country={Egypt}}

\begin{abstract}
%A procedural level generator is a tool that generates levels from noise. One approach to build generators is using machine learning, but given the training data rarity, multiple methods have been proposed to train generators from nothing. However, level generation tasks tend to have sparse feedback, which is commonly mitigated using game-specific supplemental rewards. This paper proposes a novel approach to train generators from nothing by learning at multiple level sizes starting from a small size up to the desired sizes. This approach employs the observed phenomenon that feedback is denser at smaller sizes to avoid supplemental rewards. It also presents the benefit of training generators to output levels at various sizes. We apply this approach to train controllable generators using generative flow networks. We also modify diversity sampling to be compatible with generative flow networks and to expand the expressive range. The results show that our methods can generate high-quality diverse levels for Sokoban, Zelda and Danger Dave for a variety of sizes, after only 3h 29min up to 6h 11min (depending on the game) of training on a single commodity machine. Also, the results show that our generators can output levels for sizes that were unavailable during training.
A level generator is a tool that generates game levels from noise. Training a generator without a dataset suffers from feedback sparsity, since it is unlikely to generate a playable level via random exploration. A common solution is shaped rewards, which guides the generator to achieve subgoals towards level playability, but they consume effort to design and require game-specific domain knowledge. This paper proposes a novel approach to train generators without datasets or shaped rewards by learning at multiple level sizes starting from small sizes and up to the desired sizes. The denser feedback at small sizes negates the need for shaped rewards. Additionally, the generators learn to build levels at various sizes, including sizes they were not trained for. We apply our approach to train recurrent auto-regressive generative flow networks (GFlowNets) for controllable level generation. We also adapt diversity sampling to be compatible with GFlowNets. The results show that our generators create diverse playable levels at various sizes for Sokoban, Zelda, and Danger Dave. When compared with controllable reinforcement learning level generators for Sokoban, the results show that our generators achieve better controllability and competitive diversity, while being $9\times$ faster at training and level generation.
\end{abstract}

\begin{graphicalabstract}
\includegraphics[width=\textwidth]{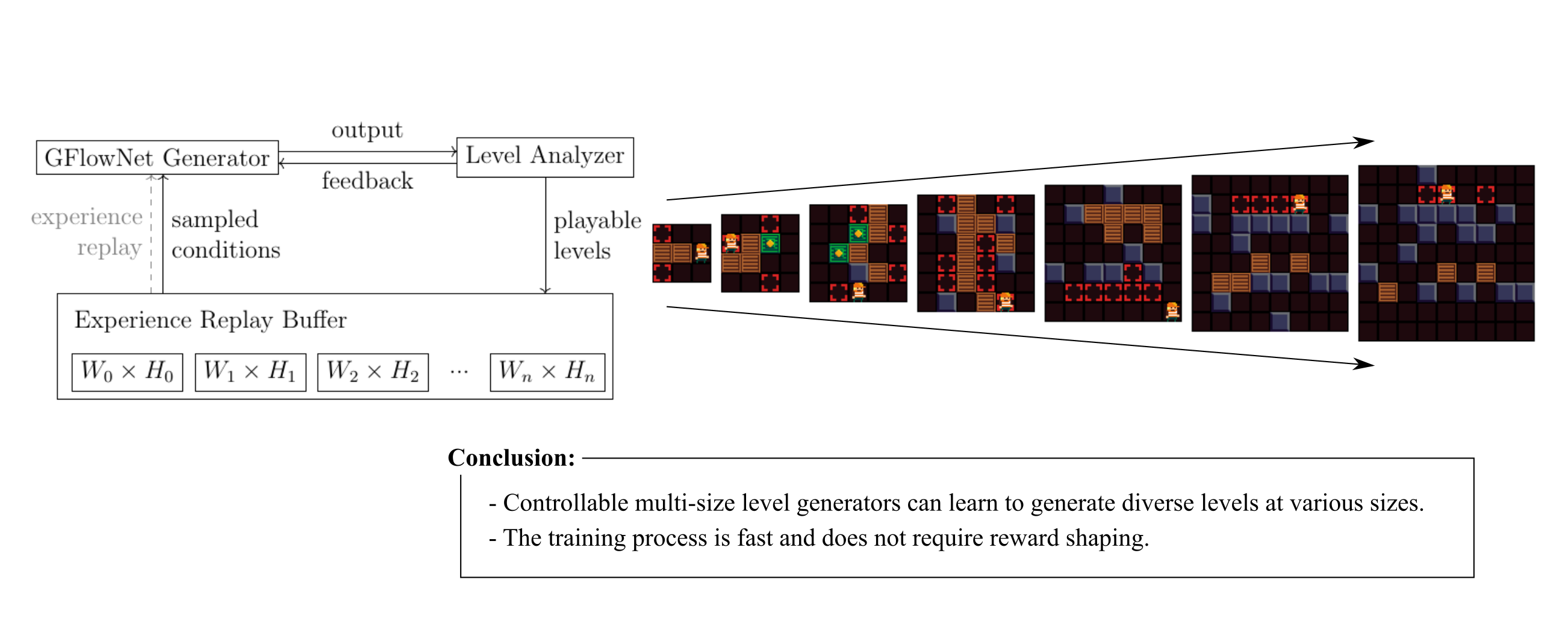}
\end{graphicalabstract}

\begin{highlights}
\item This paper proposes a novel approach to train generators without any training data or shaped rewards by learning at multiple sizes.
\item This paper presents an application of the multi-size training approach to train GFlowNets for controllable 2D tile-based game level generation at various sizes.
\item This paper modifies diversity sampling for compatibility with GFlowNets, and to expand the generators' expressive ranges.
\end{highlights}


\end{frontmatter}


\section{Introduction} \label{sec:intro}

Procedural Content Generation (PCG) is the process of generating content automatically using algorithms. PCG has been commonly used to generate levels and other assets for games since decades ago (e.g., Rogue \cite{Rogue1980} from 1980). In some games, a larger level size could allow the designer to build a more interesting environment. For example, any $3 \times 3$ Sokoban \cite{Sokoban1982} level can be solved in $14$ steps or less and they are unlikely to pose any challenge to human players. Comparatively, there are many larger Sokoban levels that demands a higher level of puzzle solving skills and long-term planning. However, training a level generator without training examples tends to be harder as the level size increases. The larger the level, the higher the chance that a bad action is taken during the generation process, rendering the level unplayable. Therefore, if the generator receives no positive feedback except for playable outputs, training becomes challenging due to the sparse feedback.

Some methods such as Procedural Content Generation via Reinforcement Learning (PCGRL) \cite{Khalifa2020PCGRL} and Neural Cellular Automata for Level Generation \cite{Earle2022NCA} mitigates the sparse feedback using shaped rewards which are designed for each game to guide the generators towards satisfying the game's functional requirements. However, reward shaping consumes effort and requires game-specific domain knowledge. We propose a novel approach where the generator starts training on smaller levels, then expands to larger ones as it improves. This approach is based on two assumptions. The first assumption is that it is more likely to generate a desirable level via random exploration if the level size is smaller, which is true for all the games we tested. The second assumption is that the knowledge gained from learning to generate small levels is useful for learning to generate larger ones, which is confirmed for our generator's architecture by the results. This approach enables efficient training without reward shaping, which simplifies adapting the generator to new games. It also produces generators that can create levels at various sizes, including sizes they were not trained for (out-of-training sizes). In this paper, we apply our approach to train recurrent auto-regressive generative flow networks (GFlowNets) \cite{Bengio2021FlowNB} for controllable level generation. By utilizing a recurrent architecture, our generators are globally consistent (make decisions that are consistent with the whole level) which is required to generate diverse playable levels for games such as Sokoban, Zelda and Danger Dave. We also adapt diversity sampling \cite{Zakaria2022plgdl} to work with GFlowNets, and modify it to expand the generators' expressive range. Finally, we present our results for three 2D tile-based games: Sokoban, Zelda and Danger Dave. The results show that our generators produce high-quality diverse levels at various sizes. When compared with Controllable PCGRL \cite{Earle2021ConPCGRL} for Sokoban, we show that our generators exhibits better controllability and competitive diversity, while being $9\times$ faster at training and level generation.

Thus, in this paper, we present the following contributions:
\begin{enumerate}
	\item Propose a novel approach to train generators without any training data or shaped rewards by learning at multiple sizes.
	\item Apply our multi-size training approach to train GFlowNets for controllable 2D tile-based game level generation at various sizes.
	\item Modify diversity sampling \cite{Zakaria2022plgdl} for compatibility with GFlowNets, and to expand the generators' expressive ranges.
	\item Present and discuss the results of the proposed methods on Sokoban, Zelda, and Danger Dave.
\end{enumerate}

This paper is organized as follows: Section \ref{sec:relwork} briefly mentions the related works, then section \ref{sec:methods} presents the proposed approach. In section \ref{sec:expsetup}, the experimental setup is detailed, then section \ref{sec:results} presents and discusses the results. Finally, section \ref{sec:conc} concludes this paper.

\section{Related Works}\label{sec:relwork}

\subsection{Procedural Level Generation}\label{ssec:plg}

Many recent works focus on learning level generators from little to no training data. If the dataset is small, different methods to bootstrap a generator using a small dataset were proposed \cite{Torrado2020Bootstrap,Siper2022PoD}. Diversity sampling \cite{Zakaria2022plgdl} was proposed to improve the solution diversity of bootstrapped Sokoban level generators. To learn a generator without training data, Procedural Content Generation via Reinforcement Learning (PCGRL) \cite{Khalifa2020PCGRL} formulates the level generation problem as a reinforcement learning (RL) task. Generative Playing Networks \cite{Bontrager2021GPN} train the generator without training data using the feedback from an RL agent, which is trained to solve the generator's output. Adversarial Reinforcement Learning for PCG \cite{Linus2021ARLPCG} follows a similar idea except that the generator is also an RL agent. Other approaches include using Quality-Diversity search to learn a diverse set of Neural Cellular Automata \cite{Earle2022NCA}, and using Neuroevolution to learn iterative level generators \cite{Beukman2022NENS}. In addition, generators can be trained to imitate other generators such Mutation Models \cite{Khalifa2022Mutation} which learn to imitate evolution.

In \cite{Beukman2022NENS}, they showed that their iterative generator can generate levels at out-of-training sizes for the games: Mario and Maze. Since their generator has a limited receptive range, it must learn to be locally consistent regardless of the level's expanse beyond its current view. This allows their generator to generalize to arbitrary sizes, despite being trained on a single size. While local consistency is sufficient for Mario and Maze, some games (e.g., Sokoban, Zelda and Danger Dave) require global consistency, and some decisions must be backed by knowledge about the whole level. For example, Sokoban, Zelda and Danger Dave generators must ensure that only one player exists in the whole level. Hence, our generators utilize a recurrent architecture, inspired by the Long Short-Term Memory used to generate Mario levels in \cite{Summerville2016MarioLSTM}, so that they can recall any necessary information about their past decisions.

\subsection{Generative Flow Networks}\label{ssec:gfn}

Generative flow networks (GFlowNets) \cite{Bengio2021FlowNB} learn to build compositional content by applying a sequence of actions. GFlowNets formulates the problem as a directed acyclic flow graph where the generation process starts at an empty object $s_0$ (the root node) and by applying an action sequence (each denoted by an edge), it reaches a complete object $s_f$ (a leaf node). Let $z_0$ denote the flow entering the graph through the root node, which would branch at nodes and flow though the edges till it pours into the leaves. The goal of GFlowNets is to learn a flow such that each leaf node receives a share equal to its reward $R(s_f)$. Then, GFlowNets can be used as a stochastic policy where the probability of each transition $P_f(s_{i+1}|s_i)$ is proportional to the flow going through its corresponding edge. So, GFlowNets has been used to sample a diverse set of compositional objects such molecular graphs \cite{Bengio2021FlowNB}, thus they are a good match for procedural level generation.

Out of the loss functions proposed for GFlowNets, we will focus on the trajectory balance loss function \cite{Malkin2022TB}, which is shown in Eq. \eqref{eq:tbloss}. It works by matching the forward and backward flow through a given trajectory. The trajectory could be sampled from the current policy or from a dataset. This loss function $L$ requires the estimated source flow $z_0$, the reward function $R(s_f)$, the forward policy $P_f(s_{i+1}|s_i)$, and the backward policy $P_b(s_i|s_{i+1})$. For auto-regressive sequence generation, each state has only one predecessor, so $P_b(s_i|s_{i+1})$ is always $1$.

\begin{equation}\label{eq:tbloss}
	L = \log(\frac{z_0 \prod_{i=1}^{f} P_f(s_i|s_{i-1})}{R(s_f) \prod_{i=1}^{f} P_b(s_{i-1}|s_i)})^2
\end{equation}

\section{Proposed Approach}\label{sec:methods}

\subsection{Multi-Size Generator Training}

The multi-size training process trains the generator at multiple sizes in parallel. The sizes include the desired sizes, a seed size, and intermediate sizes to facilitate the generator's transition from the seed size to the desired sizes. The seed size is recommended to be small (close or equal to the smallest possible size) to increase the feedback density. More seed sizes can be added if the user is unsure about the smallest possible size for their game. Initially, the generator is trained at the seed sizes only to avoid distraction by the negative feedback at large sizes. As soon as the generator yields any playable level at one of the other sizes (intermediate and desired sizes), it is added to the training sizes. So, the set of training sizes automatically expands as the generator improves. 

To generate levels at a variety of sizes, the generator's network must accommodate different level sizes without changing its architecture. So, we formulate the problem as an auto-regressive sequence generation task and use recurrent neural networks (RNN), since they can store and recall information about all the previous steps, which is crucial for global consistency. Also, RNNs can generate sequences of different lengths without any architectural changes, so they are suitable for the multi-size level generation task.

\subsection{Multi-size Generative Flow Networks}

A level generator can be trained as a generative flow network without requiring a dataset. We picked a conditional RNN and trained it using the auto-regressive trajectory balance loss function \cite{Malkin2022TB}. Therefore, the network needs to learn the source flow $z_0$ and the forward policy $P_f$. The loss function can be written as shown in Eq. \eqref{eq:ms_tbloss}.

\begin{equation}\label{eq:ms_tbloss}
	L = \biggl[ \log(z_0(u | \gamma_{(w,h)})) + \sum_{i=1}^{f} \log(P_f(s_i| s_{i-1}, u, w, h, \theta)) - \log(R(s_f|u)) \biggr]^2
\end{equation}

$L$ is the trajectory balance loss where the trajectory is $[s_0, s_1, ..., s_f]$. $u$ are the control values, $w$ \& $h$ are the level's width \& height respectively, $\gamma_{(w,h)}$ are the source flow estimation network's weights for the size $(w,h)$, $\theta$ are the forward policy RNN weights and $R(s_f|u)$ is the generated level reward given the requested controls. Unlike the forward policy network, the source flow network has different weights for each level size to improve the training stability since the training for each size progresses at a different pace. So, the model cannot estimate the source flow for out-of-training sizes, but it is not an issue since the generation process only needs the forward policy.

Since the probability of generating a level is directly proportional to its reward, it would seem intuitive that an undesirable level's reward $R^-$ should be 0. However, the loss function operates in the log space and a zero reward would introduce $-\infty$ to the gradient calculation. So, a non-zero $R^-$ value must be picked, while restricting the undesirable levels' coverage of the output distribution to be less than a certain value $P$. To calculate a bound on $R^-$, we assume that $P = 50\%$ (for simplicity, but a smaller value can be picked for a tighter bound) and the level space contains only one desirable level (the worst-case scenario). Since the level space size is $|A|^{wh}$, where $A$ is the tileset, the remaining $|A|^{wh} - 1$ levels are undesirable. So, we want to satisfy the bound in Eq. \ref{eq:reward_bound} where $R^+$ is the desirable level's reward. Assuming $R^+ \ge 1$, a valid value for $R^-$ would be $|A|^{-wh}$. Overall, the reward function in the log-space is shown in Eq. \ref{eq:ms_gfn_rew} where the $\hat{u}(l)$ are the actual values of the level properties corresponding to the controls $u$.

\begin{equation}\label{eq:reward_bound}
	R^+ > (A^{wh} - 1) R^- > 0
\end{equation}

\begin{equation}\label{eq:ms_gfn_rew}
\log(R(l|u)) =
\begin{cases}
	0 & \text{if l is playable \& } \hat{u}(l) = u \\
	-wh\log|A| & \text{otherwise}
\end{cases}
\end{equation}
	
To increase the sample efficiency, an experience replay buffer is populated with new playable levels generated during training alongside its actual control values (not the controls used to generate it). Populating a replay buffer can be seen as an equivalent to the bootstrapping method proposed in \cite{Torrado2020Bootstrap}.

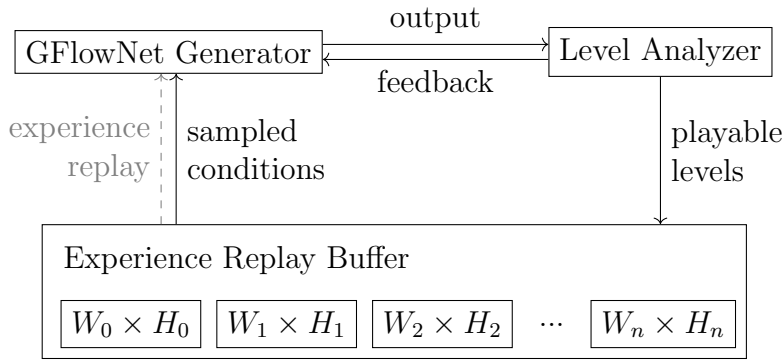
\begin{figure}[t]
	\centering
	\begin{tikzpicture}
		
		\tikzstyle{block} = [rectangle, draw=black] % style for modules
		\tikzstyle{cat} = [circle, fill=black] % style for concatenation
		\tikzstyle{hiddenline} = [dashed, draw=gray]  % style for hidden state lines
		\tikzstyle{hiddentext} = [text=gray]  % style for hidden state text
		
		\node[] (buff-title) {Experience Replay Buffer};
		\node[block, below = 0.2 of buff-title, xshift=-40.0] (s0) {$W_0\times H_0$};
		\node[block, right = 0.2 of s0] (s1) {$W_1\times H_1$};
		\node[block, right = 0.2 of s1] (s2) {$W_2\times H_2$};
		\node[right = 0.2 of s2] (sother) {...};
		\node[block, right = 0.2 of sother] (sn) {$W_n\times H_n$};
		
		\node[block, fit={(buff-title) (s0) (s1) (s2) (sother) (sn)}] (buff) {};
		
		\node[block, above = 2.0 of buff, xshift=-3.0cm] (gen) {GFlowNet Generator};
		
		\node[block, right = 3.0 of gen] (an) {Level Analyzer};
		\draw[->] ([yshift=0.1cm]gen.east) -- node [above] {output} ([yshift=0.1cm]an.west);
		\draw[->] ([yshift=-0.1cm]an.west) -- node [below] {feedback} ([yshift=-0.1cm]gen.east);
		
		\draw[->] (an.south) -- node[right, align=left]{playable \\ levels} (an.south |- buff.north);
		
		\draw[hiddenline, ->] ([xshift=-0.1cm] gen.south |- buff.north) -- node[hiddentext, left, align=right]{experience \\ replay} ([xshift=-0.1cm] gen.south);
		
		\draw[->] ([xshift=0.1cm] gen.south |- buff.north) -- node[right, align = left]{sampled \\ conditions} ([xshift=0.1cm] gen.south);
		
	\end{tikzpicture}
	\caption{The GFlowNet level generator training process. First, the generator is requested to generate output (levels) given a set of conditions sampled from the replay buffer. Then, the analyzer checks the output, send feedback (rewards) to the generator, and add any new playable levels to the replay buffer. The generator is trained using the analyzer's feedback and using levels sampled from the replay buffer (as denoted by the dashed arrow).} \label{fig:sys}
\end{figure}

Since the model is conditional, controls must be supplied to the network to roll out new trajectories, but as discussed in \cite{Zakaria2022plgdl}, the controls have no inherent parameterized distribution from which they can be sampled. In \cite{Earle2021ConPCGRL}, the controllable PCGRL environments sample the controls from a uniform distribution bounded by a user-defined range, so it assumes the user knows the desired bounds beforehand. Also, a tight bound will limit the generator's range, while a large bound may contain many regions of unsatisfiable values, thus decreasing the chances of generating playable levels during training. In \cite{Zakaria2022plgdl}, it was proposed to learn the distribution of controls from the augmented dataset. For testing the generators, it was proposed to fit a Gaussian Mixture Model (GMM) on the final augmented dataset. During training, the controls were sampled from a uniform distribution bounded by the minimum and maximum control values in the augmented dataset. It is faster to update compared to GMMs and it will expand as the training proceeds, but such an expansion could lead to sampling more unsatisfiable controls. So, we suggest sampling controls from the replay buffer, then adding some noise, to explore the regions surrounding the current generator range. To avoid bias towards the replay buffer modes, we suggest sampling conditions via diversity sampling. If the replay buffer for a certain level size is empty, the controls are sampled from the closest populated buffer based on level size ($argmin_i |wi - w| + |hi - h|$). The same criterion is applied to a pick a GMM for generating levels with out-of-training sizes. If the replay buffer is empty for all the sizes, we assign any random numbers to the conditions. To sum up the interactions between the different components, \figurename\ \ref{fig:sys} shows a block diagram of our GFlowNet level generator training process.

Since random noise is added to the sampled controls, the values could end up invalid (e.g., 2.5 crates). Even if the network may have no problem generating levels from invalid control values (usually true if they are close to the generator's range), they will cause an issue while deciding the level reward $\log(R(l|u))$. A solution is to snap the sampled controls to the nearest valid values. For example, a sampled crate count for Sokoban could be rounded to the nearest integer then clamped to the range $[1, wh - 2]$. Clamping is optional, especially when the bounds are unknown. Having a few unsatisfiable controls during training did not cause problems in our experiments.

In \cite{Zakaria2022plgdl}, the controls were level properties divided by functions of the level size as a replacement for input normalization. The same idea is applied here to improve the usability of condition models across sizes. For example, generating a level with a 250-step solution is unlikely if the level is $5 \times 5$, but more likely if it becomes $7 \times 7$. So, the GMM should not learn that a 250-step solution is unlikely. If the solution length is divided by the level area, the GMM would learn that generating a level, whose solution length is $10\times$ the level area, is unlikely regardless of the size. This presents no generalization guarantees, especially since the denominators were picked by intuition in our experiments. While trying to tune the denominators, no notable performance differences were observed when generating at in-training sizes, but they had some effect for out-of-training sizes. Still, creating a tailored GMM for the targeted size could improve the results. In that case, a two-step process can be followed: generate a level sample at the targeted size using controls sampled from the GMM for the closest in-training size, then fit a GMM using the playable portions of the generated sample. In our experiments, using the tailored GMM has always increased the probability of generating a playable level, and for some games, it also improved the controllability. But for Sokoban, the diversity dropped, since the tailored GMM was tighter than the GMM for the closest in-training size. So, we use tailored GMMs for out-of-training sizes with some games only, as will be stated in section \ref{ssec:games}.

\subsection{Diversity Sampling and Reward}

Diversity sampling was proposed \cite{Zakaria2022plgdl} to increase the solution diversity of Sokoban generators trained with bootstrapping \cite{Torrado2020Bootstrap}. Since our generator is trained without any training data, the initial replay buffer population will likely have a limited diversity, so diversity sampling is crucial for our method. When diversity sampling is applied, the levels are clustered based on their properties and training batches are collected by sampling each level uniformly from a uniformly sampled cluster. So, the probability $P_{div}$ of sampling a level $l$ is as shown in Eq. \eqref{eq:prop_div} where $C_{key(l)}$ is the cluster containing $l$ and $N$ is the number of clusters. When auto-regressive models are trained using the cross-entropy loss, the output distribution follows the training data distribution. In GFlowNets, however, the output distribution follows the reward distribution. So, we add a diversity reward $R_{div} \propto P_{div}$. Since the reward function in Eq. \eqref{eq:ms_gfn_rew} requires that $R^+ \ge 1$, the diversity reward is picked to be as shown in Eq. \eqref{eq:rew_div} where $\{C1, ..., Cn\}$ are the clusters. Despite the diversity reward, diversity sampling is still needed to ensure that rare levels appear frequently during training.

\begin{equation}\label{eq:prop_div}
P_{div}(l) = \frac{1}{N|C_{key(l)}|}
\end{equation}

\begin{equation}\label{eq:rew_div}
R_{div}(l) = \frac{\max_i|C_i|}{|C_{key(l)}|}
\end{equation}

In \cite{Zakaria2022plgdl}, the levels were clustered based on a distilled form of the solution, called the solution signature. However, the solution signature proposed in \cite{Zakaria2022plgdl} is only applicable to Sokoban. And, as the results will show, the levels could have a large variety of signatures while only covering a small range over the pushed crate count and the solution length. So, we changed the clustering key to be a tuple containing a set of properties. For example, we picked a tuple of the pushed crate count and the solution length as the cluster key for Sokoban. Additionally, a granularity can be picked to group together close property values. For example, we divided the solution length by $(w + h)$ then floored it before adding it to the cluster key, so that minor solution length differences are ignored. Thus, a level with a unique solution length would not be considered unique, if it is in the vicinity of levels with the same pushed crate count and close solution lengths. Deciding the cluster key properties and granularities is left for the users to pick based on their preferences.

An optional addition is the property reward $R_{prop}$ where the generator is rewarded based on the generated level's properties. For example, if difficult levels are more desirable, a reward proportional to the level difficulty could be added. Overall, the reward function $\log(R_{total}(l|u))$ used in our experiments is as shown in Eq. \eqref{eq:total_ms_gfn_rew}. It is noteworthy that the diversity and the property rewards are added even when the generated level does not satisfy the controls. So, a rare and/or a generally desirable level is still less preferred to a level that satisfies the controls, but more preferred to other more common levels.

\begin{equation}\label{eq:total_ms_gfn_rew}
	\begin{aligned}
		\log(R_{div}(l)) =
		\begin{cases}
			\log(max_i |C_i|) - \log|C_{key(l)}| & \text{if $l$ is playable} \\
			0 & \text{otherwise}
		\end{cases} \\
		\log(R_{total}(l|u)) = \log(R(l|u)) + \log(R_{div}(l)) + \log(R_{prop}(l))
	\end{aligned}
\end{equation}
	
A remaining issue is that obvious tile patterns may appear in a significant portion of the generated levels, despite containing a diverse variety of solutions and properties. This issue can be solved using data augmentation where the levels are randomly flipped (vertically and/or horizontally depending on the game rules) during training.

\section{Experimental Setup}\label{sec:expsetup}

\subsection{Model Architecture and Hyperparameters}\label{ssec:arch_hyperparam}

\begin{figure}[t]
	\centering
	\begin{tikzpicture}
		
		\tikzstyle{layer} = [rectangle, draw=black] % style for modules
		\tikzstyle{cat} = [circle, fill=black] % style for concatenation
		\tikzstyle{hiddenline} = [dashed, draw=gray]  % style for hidden state lines
		\tikzstyle{hiddentext} = [text=gray]  % style for hidden state text
		
		%%%% The GRUs
		
		% The concat node combining the inputs
		\node[cat] (cat0) {};
		
		% The 1st GRU Cell
		\node[layer, above=1 of cat0] (gru1) {GRU Cell};
		% The line connecting GRU1 to its input
		\draw[->] (cat0) -- node[anchor=north west, at end] {input} (gru1);
		
		% The hidden i/o of GRU1 and their arrows
		\node[left = 0.75 of gru1, hiddentext] (h1p) {$h^1_{i-1}$};
		\draw[->, hiddenline] (h1p) -- (gru1);
		\node[right = 0.75 of gru1, hiddentext] (h1n) {$h^1_i$};
		\draw[->, hiddenline] (gru1) -- (h1n);
		
		% The concat node after GRU1
		\node[cat, above=0.5 of gru1] (cat1) {};
		% The input for the concat1 coming from the 1st concat and the 1st GRU
		\draw[->] (cat0.north) |- +(-1.25,0.5) |- (cat1.west);
		\draw[->] (gru1.east) -- +(0.25,0) |- (cat1.east);
		
		% The 2nd GRU Cell
		\node[layer, above=0.5 of cat1] (gru2) {GRU Cell};
		\draw[->] (cat1) -- node[anchor=north west, at end] {input} (gru2);
		
		% The hidden i/o of GRU2 and their arrows
		\node[left = 0.75 of gru2, hiddentext] (h2p) {$h^2_{i-1}$};
		\draw[->, hiddenline] (h2p) -- (gru2);
		\node[right = 0.75 of gru2, hiddentext] (h2n) {$h^2_i$};
		\draw[->, hiddenline] (gru2) -- (h2n);
		
		% The concat node after GRU1
		\node[cat, above=0.5 of gru2] (cat2) {};
		% The input for the concat2 coming from the 1st concat and the two GRUs
		\draw[->] (cat0.north) |- +(-1.25,0.5) |- (cat2.west);
		\draw[->] (gru2.east) -| +(0.25,0.5) -| (cat2.south);
		\draw[->] (gru1.east) -- +(0.5,0) |- (cat2.east);
		
		%%%% The Action Module
		
		% The action module, its input & output and their arrows
		\node[layer, above=0.5 of cat2] (act) {Action Module};
		\draw[->] (cat2) -- (act);
		\node[above=0.5 of act] (pact) {$P(t_i)$};
		\draw[->] (act) -- (pact);
		
		%%%% The inputs
		
		% The previous tile as an input
		\node[left=0.5 of cat0, anchor=east] (tile) {$t_{i-1}$};
		\draw[->] (tile) -- (cat0);
		% The new row flag as an input
		\node[right=0.5 of cat0, anchor=west] (flag) {$row_{i-1} \neq row_i$};
		\draw[->] (flag) -- (cat0);
		
		%%%% The Conditonal Embedding Module
		
		% The conditional embeding module and an arrow from it to the input concat
		\node[layer, below=0.5 of cat0, align=center] (cemod) {Conditional \\ Embedding \\ Module};
		\draw[->] (cemod) -- (cat0);
		
		% The concat before the CE module + the arrow to CE module
		\node[cat, below=0.25 of cemod] (catce) {};
		\draw[->] (catce) -- (cemod);
		
		% The Controls input + arrow to CE concat
		\node[left=0.5 of catce, anchor=east] (controls) {Controls};
		\draw[->] (controls) -> (catce);
		
		% The Level size input + arrow to CE concat
		\node[right=0.5 of catce, anchor=west] (size) {Level Size};
		\draw[->] (size) -> (catce);
		
	\end{tikzpicture}
	\caption{The Generator's Network Architecture. Black circles represent concatenation operators. $t_i$ is the tile to generate at step $i$, and $row_i$ is the row index of the tile $t_i$.} \label{fig:netarch}
\end{figure}
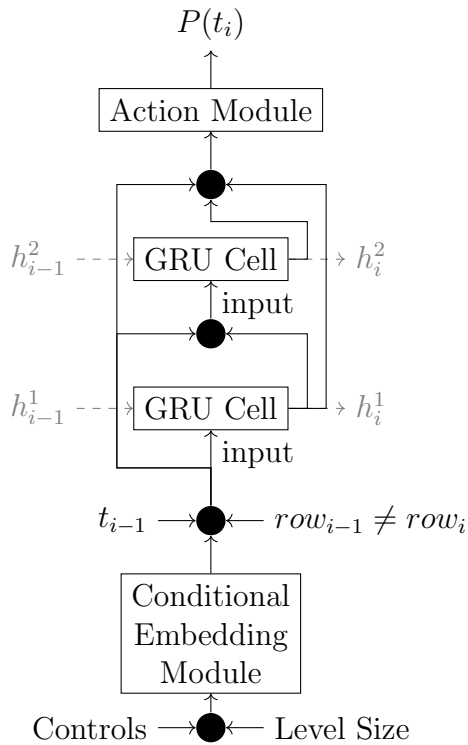

All the experiments use the model architecture shown in \figurename\ \ref{fig:netarch}. The hidden size of each GRU cell is $128$. The conditional embedding module consists of 2 feed-forward layers: 16 neurons (Leaky ReLU) and 32 neurons (No activation). The action module also consists of 2 feed-forward layers: 32 neurons (Leaky ReLU) and $|A|$ neurons (Softmax) where A is the tileset. The last layer's weights and biases are initialized to zero. A source flow estimator is instantiated for every level size, each consisting of 2 feed-forward layers: 32 neurons (Leaky ReLU) and 1 neuron (No activation). The last layer's weights and bias are initialized to $0$ and the output is treated as $\log(z_0)$. The model traverses the level in a row-wise snake-like pattern (each row is traversed in the reverse order of the previous row). The models are trained using RMSProp \cite{Tieleman2012RMSPROP} with a learning rate of $0.001$ except for the source flow estimators whose learning rate is $0.01$. The batch size is $32$. All the experiments are run for $10,000$ iterations. After training, we follow \cite{Zakaria2022plgdl} and fit a GMM to the final experience replay buffer, and sample the controls from it during testing. We use a Bayesian Gaussian Mixture Model \cite{Roberts1998BGMM} with 16 components fit for 100 iterations.

To examine a generator's output quality and diversity, $10,000$ levels are generated where the controls are sampled unconditionally from the GMM. To examine a generator's controllability, a level sample is generated for each control where the tested control is fixed, and the others are sampled conditionally from the GMM. The sample size and the tested control values will be stated as we define each control in section \ref{ssec:games}. The test setup for Sokoban matches the setup in \cite{Zakaria2022plgdl} to facilitate comparing our method with the methods included in their study. To test the training stability, we run each experiment 5 times, compute the results for each run separately, then report the mean and the standard deviation. All the aforementioned experiments are run on an 8-core 3.7 GHz CPU (16-threads) and Nvidia RTX 3070 GPU laptop running Windows 11.

To put our results in context, we compare them with Controllable PCGRL \cite{Earle2021ConPCGRL} for generating Sokoban levels at the size $7\times 7$. We run the Controllable PCGRL experiment 5 times using the training and generation setup defined in \cite{Zakaria2022plgdl} with the exception of the control bounds, which were expanded to be $[1, 14]$ for the pushed crates, and $[1, 343]$ for the solution length, to better match the range of our generators. We picked Controllable PCGRL for our comparison, since it presents the closest set of features by being controllable and trainable without a dataset. In addition, Controllable PCGRL exhibits high quality and diversity for Sokoban level generation as shown in \cite{Zakaria2022plgdl}. The experiments were run on a 16-core 3.4 GHz CPU (32-threads) and Nvidia RTX 3090 GPU desktop running Ubuntu 20.04. To compare the training and generation times, we also run 5 experiments of our generator with diversity sampling, property reward and data augmentation for $20,000$ iterations on the same machine.

\subsection{Games}\label{ssec:games}

The methods are tested on Sokoban, Zelda and Danger Dave. In this subsection, we briefly introduce the three games, their functional requirements, the in-training \& out-of-training level sizes, the controls, the cluster key and the property reward. For each control, we explain it, state its denominator $den$ (used to divide the property when supplied to the network and the GMM fitter), the noise $z_s$ added after sampling during training, and the test sample defined by the sampled level count $n_{test}$ for each control value in the set $C_{test}$. $w$ and $h$ are used to denote the level's width and height. For some controls, the noise has a positive mean to incentivize exploring towards a desired direction. The tilesets of the three games are shown in \tablename\ \ref{tbl:tileset}.

\newcommand{\lowincludegraphics}[2][]{%
	\raisebox{-0.7\dp\strutbox}{\includegraphics[#1]{#2}}%
}

\begin{table}
	\caption{The Tilesets of the 3 games included in the experiments.}\label{tbl:tileset}
	\centering
	\begin{tabular}{cl|cl|cl}
		\toprule
		\multicolumn{2}{c|}{Sokoban} & 
		\multicolumn{2}{c}{Zelda} & 
		\multicolumn{2}{|c}{Danger Dave} \\
		\midrule
		
		\lowincludegraphics{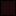} & Empty & \lowincludegraphics{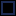} & Empty & \lowincludegraphics{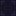} & Empty \\ 
		
		\lowincludegraphics{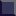} & Wall & \lowincludegraphics{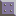} & Wall & \lowincludegraphics{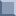} & Wall \\ 
		
		\lowincludegraphics{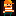} & Player & \lowincludegraphics{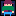} & Player & \lowincludegraphics{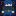} & Player \\ 
		
		\lowincludegraphics{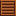} & Crate & \lowincludegraphics{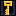} & Key & \lowincludegraphics{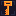} & Key \\ 
		
		\lowincludegraphics{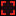} & Goal & \lowincludegraphics{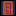} & Door & \lowincludegraphics{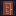} & Door \\ 
		
		\lowincludegraphics{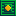} & Crate on Goal & \lowincludegraphics{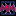} & Bat & \lowincludegraphics{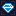} & Diamond \\ 
		
		\lowincludegraphics{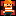} & Player on Goal & \lowincludegraphics{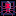} & Spider & \lowincludegraphics{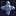} & Spike \\

		&& \lowincludegraphics{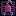} & Scorpion && \\
		\bottomrule
	\end{tabular}
\end{table}

\subsubsection{Sokoban}

\begin{figure*}[!t]
	\centering
	
	\begin{subfigure}{0.15\columnwidth}
		\centering
		\includegraphics[width=\columnwidth]{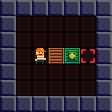}
		\caption{$t=0$}
	\end{subfigure}~
	\begin{subfigure}{0.15\columnwidth}
		\centering
		\includegraphics[width=\columnwidth]{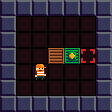}
		\caption{$t=1$}
	\end{subfigure}~	
	\begin{subfigure}{0.15\columnwidth}
		\centering
		\includegraphics[width=\columnwidth]{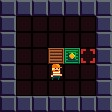}
		\caption{$t=2$}
	\end{subfigure}~
	\begin{subfigure}{0.15\columnwidth}
		\centering
		\includegraphics[width=\columnwidth]{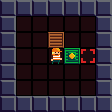}
		\caption{$t=3$}
	\end{subfigure}~
	\begin{subfigure}{0.15\columnwidth}
		\centering
		\includegraphics[width=\columnwidth]{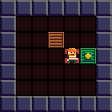}
		\caption{$t=4$}
	\end{subfigure}~
	\begin{subfigure}{0.15\columnwidth}
		\centering
		\includegraphics[width=\columnwidth]{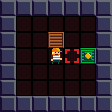}
		\caption{$t=5$}
	\end{subfigure}
	
	\begin{subfigure}{0.15\columnwidth}
		\centering
		\includegraphics[width=\columnwidth]{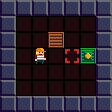}
		\caption{$t=6$}
	\end{subfigure}~
	\begin{subfigure}{0.15\columnwidth}
		\centering
		\includegraphics[width=\columnwidth]{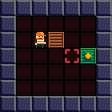}
		\caption{$t=7$}
	\end{subfigure}~	
	\begin{subfigure}{0.15\columnwidth}
		\centering
		\includegraphics[width=\columnwidth]{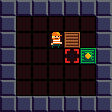}
		\caption{$t=8$}
	\end{subfigure}~
	\begin{subfigure}{0.15\columnwidth}
		\centering
		\includegraphics[width=\columnwidth]{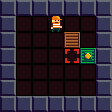}
		\caption{$t=9$}
	\end{subfigure}~
	\begin{subfigure}{0.15\columnwidth}
		\centering
		\includegraphics[width=\columnwidth]{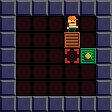}
		\caption{$t=10$}
	\end{subfigure}~
	\begin{subfigure}{0.15\columnwidth}
		\centering
		\includegraphics[width=\columnwidth]{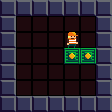}
		\caption{$t=11$}
	\end{subfigure}
	
	\caption{A Sokoban example level and a sequence of steps to solve it.}
	\label{fig:sokoban_example}
\end{figure*}

Sokoban \cite{Sokoban1982} is a top-down puzzle game where the player pushes crates around the level until each crate is located on a goal tile. The player can only push one crate at a time and cannot push a crate into a wall or another crate. The functional requirements are: there is only one player, the crate and goal counts are equal, at least one crate is not on a goal, and the solver can win the level. \figurename\ \ref{fig:sokoban_example} shows a Sokoban level example and a possible solution for it. The generator is trained on the sizes $3\times 3$, $4\times 4$, $5\times 5$, $6\times 6$ and $7\times 7$. During testing, the sizes $8\times 8$ and $9\times 9$ are added. The solver uses Breadth-First Search with an iteration limit of $5\times 10^2$ for $3\times 3$ levels, $5\times 10^4$ for $4\times 4$, $5\times 10^5$ for $5\times 5$ and $10^6$ for the larger sizes. For data augmentation, the levels can be flipped vertically, horizontally or both. The controls are as follows:
\begin{enumerate}
	\item Pushed Crates: the number of crates that the solver pushed to solve the level. Since the generator can add a crate and a goal on the same tile, some crates may not need to be moved solve the level. These crates are not counted since they are easy to add and remove without affecting the level solution. $den = (w + h)/2$, $z_s \sim U[-1, 1]$ crates, $n_{test} = 1,000$ and $C_{test} = [1, 10]$.
	\item Solution Length: the length of the shortest solution. $den = wh$, $z_s \sim U[-5, 10]$ steps, $n_{test} = 100$ and $C_{test} = [1, 100]$.
\end{enumerate}

The cluster key consists of the pushed crate count and the solution length with a granularity of $1$ and $(w + h)$ respectively. The property reward is the logarithm of the solution length. Controls for out-of-training sizes will be sampled from the GMM of the closest in-training size, since using tailored GMMs significantly decreased the diversity.

\subsubsection{Zelda}

\begin{figure*}[!t]
	\centering
	
	\begin{subfigure}{0.15\columnwidth}
		\centering
		\includegraphics[width=\columnwidth]{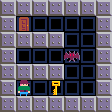}
		\caption{$t=0$}
	\end{subfigure}~
	\begin{subfigure}{0.15\columnwidth}
		\centering
		\includegraphics[width=\columnwidth]{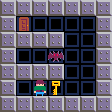}
		\caption{$t=1$}
	\end{subfigure}~	
	\begin{subfigure}{0.15\columnwidth}
		\centering
		\includegraphics[width=\columnwidth]{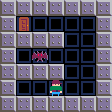}
		\caption{$t=2$}
	\end{subfigure}~
	\begin{subfigure}{0.15\columnwidth}
		\centering
		\includegraphics[width=\columnwidth]{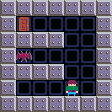}
		\caption{$t=3$}
	\end{subfigure}~
	\begin{subfigure}{0.15\columnwidth}
		\centering
		\includegraphics[width=\columnwidth]{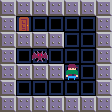}
		\caption{$t=4$}
	\end{subfigure}~
	\begin{subfigure}{0.15\columnwidth}
		\centering
		\includegraphics[width=\columnwidth]{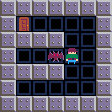}
		\caption{$t=5$}
	\end{subfigure}
	
	\begin{subfigure}{0.15\columnwidth}
		\centering
		\includegraphics[width=\columnwidth]{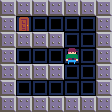}
		\caption{$t=6$}
	\end{subfigure}~
	\begin{subfigure}{0.15\columnwidth}
		\centering
		\includegraphics[width=\columnwidth]{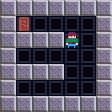}
		\caption{$t=7$}
	\end{subfigure}~	
	\begin{subfigure}{0.15\columnwidth}
		\centering
		\includegraphics[width=\columnwidth]{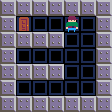}
		\caption{$t=8$}
	\end{subfigure}~
	\begin{subfigure}{0.15\columnwidth}
		\centering
		\includegraphics[width=\columnwidth]{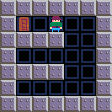}
		\caption{$t=9$}
	\end{subfigure}~
	\begin{subfigure}{0.15\columnwidth}
		\centering
		\includegraphics[width=\columnwidth]{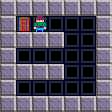}
		\caption{$t=10$}
	\end{subfigure}~
	\begin{subfigure}{0.15\columnwidth}
		\centering
		\includegraphics[width=\columnwidth]{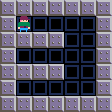}
		\caption{$t=11$}
	\end{subfigure}
	
	\caption{A Zelda example level and a sequence of steps to solve it. At time step 5, the player attacks the bat to its left.}
	\label{fig:zelda_example}
\end{figure*}

The GVGAI \cite{Liebana2019GVGAI} version of Zelda \cite{Zelda1986} is a top-down turn-based adventure game where the player has to fetch a key, then exit via the locked door. The level contains enemies that kill the player if they enter the player's tile. The player can attack an adjacent tile to kill the enemy in it. The functional requirements are: there is only one player, key \& door, the enemy count is in $[1, max(w, h)]$, there is a path from the player to the key then to the door, and every enemy in the level can reach the player. \figurename\ \ref{fig:zelda_example} shows a Zelda level example and a possible solution for it. The generator is trained on the sizes $3\times 4$, $3\times 6$, $5\times 4$, $5\times 6$, $7\times 6$, $5\times 11$ and $7\times 11$. During testing, the sizes $6\times 10$, $10\times 6$, $8\times 12$ and $9\times 13$ are added. For data augmentation, the levels can be flipped vertically, horizontally or both. The controls are as follows:
\begin{enumerate}
	\item Nearest Enemy Distance: the shortest path length to the nearest enemy. $den = wh$, $z_s \sim U[-2, 5]$ steps, $n_{test} = 100$ and $C_{test} = [1, \lfloor wh/2 \rfloor ]$.
	\item Path Length: the shortest path length to the key then to the door. $den = wh$, $z_s \sim U[-5, 10]$ steps, $n_{test} = 100$ and $C_{test} = [2, wh]$.
	\item Enemy Count: the number of enemies in the level. $den = wh$, $z_s \sim U[-1, 2]$ enemies, $n_{test} = 1,000$ and $C_{test} = [1, max(w,h)]$.
\end{enumerate}

The cluster key consists of the nearest enemy distance and the path length with a granularity of $\lfloor(w + h)/2\rfloor$ for both. The property reward is the logarithm of the path length. Tailored GMMs will be created for out-of-training sizes.

\subsubsection{Danger Dave}

\begin{figure*}[!t]
	\centering
	
	\begin{subfigure}{0.15\columnwidth}
		\centering
		\includegraphics[width=\columnwidth]{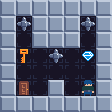}
		\caption{$t=0$}
	\end{subfigure}~
	\begin{subfigure}{0.15\columnwidth}
		\centering
		\includegraphics[width=\columnwidth]{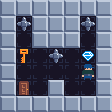}
		\caption{$t=1$}
	\end{subfigure}~	
	\begin{subfigure}{0.15\columnwidth}
		\centering
		\includegraphics[width=\columnwidth]{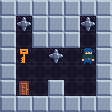}
		\caption{$t=2$}
	\end{subfigure}~
	\begin{subfigure}{0.15\columnwidth}
		\centering
		\includegraphics[width=\columnwidth]{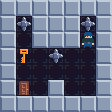}
		\caption{$t=3$}
	\end{subfigure}~
	\begin{subfigure}{0.15\columnwidth}
		\centering
		\includegraphics[width=\columnwidth]{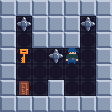}
		\caption{$t=4$}
	\end{subfigure}~
	\begin{subfigure}{0.15\columnwidth}
		\centering
		\includegraphics[width=\columnwidth]{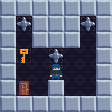}
		\caption{$t=5$}
	\end{subfigure}
	
	\begin{subfigure}{0.15\columnwidth}
		\centering
		\includegraphics[width=\columnwidth]{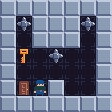}
		\caption{$t=6$}
	\end{subfigure}~
	\begin{subfigure}{0.15\columnwidth}
		\centering
		\includegraphics[width=\columnwidth]{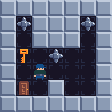}
		\caption{$t=7$}
	\end{subfigure}~	
	\begin{subfigure}{0.15\columnwidth}
		\centering
		\includegraphics[width=\columnwidth]{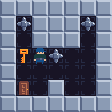}
		\caption{$t=8$}
	\end{subfigure}~
	\begin{subfigure}{0.15\columnwidth}
		\centering
		\includegraphics[width=\columnwidth]{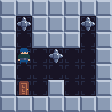}
		\caption{$t=9$}
	\end{subfigure}~
	\begin{subfigure}{0.15\columnwidth}
		\centering
		\includegraphics[width=\columnwidth]{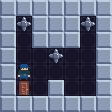}
		\caption{$t=10$}
	\end{subfigure}~
	\begin{subfigure}{0.15\columnwidth}
		\centering
		\includegraphics[width=\columnwidth]{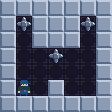}
		\caption{$t=11$}
	\end{subfigure}
	
	\caption{A Danger Dave example level and a sequence of steps to solve it. At time steps 0 and 6, the player performs a jump action.}
	\label{fig:dave_example}
\end{figure*}

Danger Dave \cite{DDave1988} is a platformer game where the player has to fetch a key, then exit through the locked door. Unlike Zelda, this game has a side view, so the player is pulled down by gravity. The level has spikes which kills the player upon touch. The level also contains diamonds which can be optionally collected. The player can move sideways and jump. The functional requirements are: there is only one player, key and door, the player is initially on a ground, the spike count is less than $(w - 1)\lfloor h / 2\rfloor$, the diamond count is in $[1, max(w, h)]$, all the diamonds are reachable (but similar to the original game, some diamonds may be located at positions where the player would die or get stuck after reaching them), and the level can be solved. \figurename\ \ref{fig:dave_example} shows a Danger Dave level example and a possible solution for it. We use the same sizes as Zelda. For data augmentation, the levels can only be flipped horizontally. The controls are as follows:
\begin{enumerate}
\item Solution Length: the minimum step count needed to solve the level. $den = wh$, $z_s \sim U[-5, 10]$ steps, $n_{test} = 100$ and $C_{test} = [2, wh]$.
\item Jump Count: the number of jumps in the solver's solution. $den = max(w, h)$, $z_s \sim U[-1, 2]$ jumps, $n_{test} = 100$ and $C_{test} = [1, \lfloor wh/4 \rfloor ]$.
\item Spike Count: the number of spikes in the levels. $den = wh$, $z_s \sim U[-1, 1]$ spikes, $n_{test} = 1,000$ and $C_{test} = [1, max(w, h)]$.
\end{enumerate}

The cluster key consists of the jump count and the solution length with a granularity of $1$ and $\lfloor(w + h)/2\rfloor$ respectively. No property reward was used for this game's experiments. Tailored GMMs will be created for out-of-training sizes.

\section{Results and Discussion}\label{sec:results}

This section presents and discusses the experimental results. The results explore the method performance regarding the output quality, diversity and controllability. Then, the training and inference times are reported. Finally, we compare our results with Controllable PCGRL.

\subsection{Quality and Diversity}\label{ssec:quality_diversity}

\begin{table}[t]
	\tiny
	\caption{Quality and Diversity of the Sokoban Generators. DS: Diversity Sampling. PR: Property Reward. AUG: Data Augmentation. `Sig.' is an abbreviation for the solution signature. Out-of-training sizes are underlined.}
	\label{tbl:sokoban_qd}
	\centering

\begin{tabular}{@{\hspace{0.1cm}} c @{\hspace{0.1cm}} c @{\hspace{0.1cm}} c @{\hspace{0.2cm}} r @{\hspace{0.2cm}} r @{\hspace{0.2cm}} r @{\hspace{0.2cm}} r @{\hspace{0.2cm}} r @{\hspace{0.2cm}} r @{\hspace{0.1cm}} }
	\toprule
	DS & PR & AUG & &\multicolumn{1}{c}{$5\times 5$} & \multicolumn{1}{c}{$6\times 6$} & \multicolumn{1}{c}{$7\times 7$} & \multicolumn{1}{c}{$\underline{8\times 8}$} & \multicolumn{1}{c}{$\underline{9\times 9}$}\\
	\midrule
	\multirow{4}{*}{. }&\multirow{4}{*}{ . }&\multirow{4}{*}{ .} & Playable\% & $\pmb{89.1\% \pm 11.2\%}$ & $\pmb{88.7\% \pm 11.6\%}$ & $\pmb{88.7\% \pm 11.7\%}$ & $\pmb{71.0\% \pm 17.6\%}$ & $\pmb{53.4\% \pm 27.7\%}$\\
	& &  & Diversity & $\pmb{0.64 \pm 0.11}$ & $0.59 \pm 0.16$ & $0.55 \pm 0.19$ & $0.25 \pm 0.56$ & $0.25 \pm 0.51$\\
	& &  & Unique Sig. & $2.2\% \pm 3.0\%$ & $2.2\% \pm 3.0\%$ & $1.5\% \pm 2.2\%$ & $1.4\% \pm 2.3\%$ & $2.0\% \pm 3.2\%$\\
	& &  & Sol. Length & $4.63 \pm 5.20$ & $5.14 \pm 5.69$ & $5.39 \pm 5.74$ & $5.64 \pm 6.56$ & $6.13 \pm 7.06$\\
	\midrule
	\multirow{4}{*}{Sig. }&\multirow{4}{*}{ . }&\multirow{4}{*}{ .} & Playable\% & $78.2\% \pm 6.5\%$ & $71.4\% \pm 5.6\%$ & $67.5\% \pm 5.0\%$ & $50.6\% \pm 4.3\%$ & $39.5\% \pm 11.3\%$\\
	& &  & Diversity & $0.58 \pm 0.04$ & $0.54 \pm 0.03$ & $0.47 \pm 0.04$ & $0.44 \pm 0.04$ & $0.42 \pm 0.05$\\
	& &  & Unique Sig. & $\pmb{82.3\% \pm 3.2\%}$ & $\pmb{69.3\% \pm 8.3\%}$ & $50.9\% \pm 15.2\%$ & $\pmb{48.2\% \pm 20.1\%}$ & $\pmb{42.1\% \pm 15.9\%}$\\
	& &  & Sol. Length & $25.68 \pm 7.08$ & $26.13 \pm 7.85$ & $25.13 \pm 8.22$ & $26.43 \pm 9.10$ & $\pmb{26.77 \pm 9.25}$\\
	\midrule
	\multirow{4}{*}{\checkmark }&\multirow{4}{*}{ . }&\multirow{4}{*}{ .} & Playable\% & $66.8\% \pm 2.8\%$ & $65.5\% \pm 4.1\%$ & $65.5\% \pm 4.8\%$ & $25.6\% \pm 11.9\%$ & $16.1\% \pm 4.8\%$\\
	& &  & Diversity & $0.55 \pm 0.07$ & $0.51 \pm 0.07$ & $0.45 \pm 0.11$ & $0.43 \pm 0.13$ & $0.43 \pm 0.12$\\
	& &  & Unique Sig. & $41.3\% \pm 15.9\%$ & $40.8\% \pm 14.9\%$ & $30.5\% \pm 9.8\%$ & $30.3\% \pm 12.3\%$ & $27.8\% \pm 10.0\%$\\
	& &  & Sol. Length & $31.09 \pm 23.93$ & $50.13 \pm 42.38$ & $70.16 \pm 50.44$ & $26.38 \pm 20.92$ & $24.28 \pm 17.40$\\
	\midrule
	\multirow{4}{*}{\checkmark }&\multirow{4}{*}{ \checkmark }&\multirow{4}{*}{ .} & Playable\% & $66.9\% \pm 4.0\%$ & $62.4\% \pm 3.6\%$ & $65.9\% \pm 5.9\%$ & $27.7\% \pm 14.3\%$ & $20.1\% \pm 13.0\%$\\
	& &  & Diversity & $0.53 \pm 0.02$ & $0.48 \pm 0.08$ & $0.41 \pm 0.06$ & $0.38 \pm 0.04$ & $0.36 \pm 0.05$\\
	& &  & Unique Sig. & $48.4\% \pm 14.1\%$ & $45.4\% \pm 12.8\%$ & $33.0\% \pm 10.4\%$ & $27.7\% \pm 10.5\%$ & $23.8\% \pm 11.8\%$\\
	& &  & Sol. Length & $\pmb{36.51 \pm 26.29}$ & $\pmb{60.50 \pm 45.68}$ & $\pmb{84.77 \pm 58.20}$ & $24.80 \pm 12.57$ & $22.78 \pm 10.76$\\
	\midrule
	\multirow{4}{*}{\checkmark }&\multirow{4}{*}{ . }&\multirow{4}{*}{ \checkmark} & Playable\% & $56.5\% \pm 6.9\%$ & $51.4\% \pm 5.6\%$ & $52.3\% \pm 4.3\%$ & $33.4\% \pm 6.4\%$ & $24.3\% \pm 7.1\%$\\
	& &  & Diversity & $0.61 \pm 0.02$ & $\pmb{0.61 \pm 0.02}$ & $\pmb{0.58 \pm 0.06}$ & $\pmb{0.52 \pm 0.06}$ & $\pmb{0.46 \pm 0.07}$\\
	& &  & Unique Sig. & $59.8\% \pm 5.4\%$ & $50.5\% \pm 7.5\%$ & $43.5\% \pm 6.9\%$ & $32.9\% \pm 7.1\%$ & $26.2\% \pm 6.7\%$\\
	& &  & Sol. Length & $28.63 \pm 18.65$ & $33.16 \pm 36.17$ & $31.95 \pm 33.46$ & $21.23 \pm 12.00$ & $20.26 \pm 9.95$\\
	\midrule
	\multirow{4}{*}{\checkmark }&\multirow{4}{*}{ \checkmark }&\multirow{4}{*}{ \checkmark} & Playable\% & $61.2\% \pm 2.4\%$ & $55.0\% \pm 3.8\%$ & $51.9\% \pm 5.7\%$ & $31.8\% \pm 3.6\%$ & $20.9\% \pm 5.7\%$\\
	& &  & Diversity & $0.59 \pm 0.01$ & $0.55 \pm 0.03$ & $0.51 \pm 0.04$ & $0.44 \pm 0.04$ & $0.40 \pm 0.05$\\
	& &  & Unique Sig. & $60.2\% \pm 2.7\%$ & $54.1\% \pm 2.3\%$ & $\pmb{51.4\% \pm 5.6\%}$ & $38.2\% \pm 3.7\%$ & $31.4\% \pm 6.5\%$\\
	& &  & Sol. Length & $30.70 \pm 19.89$ & $36.34 \pm 31.95$ & $50.12 \pm 54.49$ & $\pmb{26.65 \pm 17.20}$ & $24.87 \pm 14.19$\\
	\bottomrule
\end{tabular}
	
\end{table}

\tablename\ \ref{tbl:sokoban_qd}, \tablename\ \ref{tbl:zelda_qd} and \tablename\ \ref{tbl:dave_qd} present the results of the Sokoban, Zelda and Danger Dave generators, respectively, trained with a variety of configurations. The tables contain the following metrics:
\begin{itemize}
	\item `Playable\%' denotes the playability which is the percentage of generated levels that satisfy the functional requirements.
	\item `Diversity' is the tile-diversity which is the average tile-wise hamming distance between all pairs of playable levels divided by the level area. 
	\item `Unique Sig.' is $100\%\ \times$ the unique signatures count $/$ the playable levels count. This is available for Sokoban only.
	\item `Solution Length' presents the mean and standard deviation of the solution lengths of all the unique levels extracted from the $3$ runs. To save space, this is reported for Sokoban only.
\end{itemize}
DS, PR \& AUG stands for Diversity Sampling, Property Reward and Data Augmentation respectively. The Sokoban experimental results include two variations of diversity sampling: one using the solution signature as the key (Signature Key for short) which is marked by `Sig.' in the table, and one using a tuple (as defined in section \ref{ssec:games}) as the key (Tuple Key for short). Sizes $3 \times 3$ and $4 \times 4$ were omitted from the results to save space.

\begin{figure}[!t]
	\centering
	
	\begin{subfigure}{0.5\columnwidth}
		\centering
		\includegraphics[width=\columnwidth]{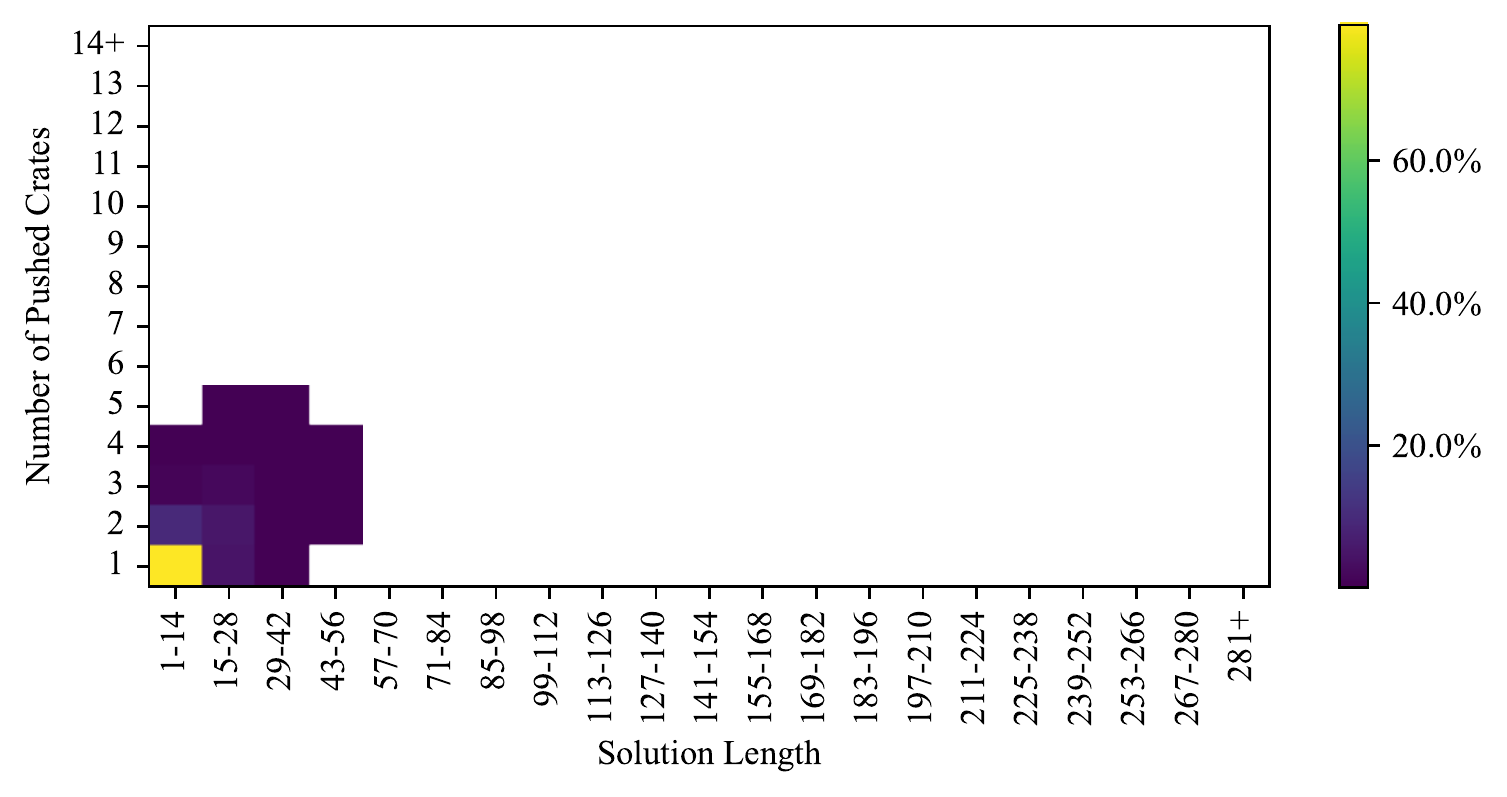}
		\caption{No DS}
	\end{subfigure}~
	\begin{subfigure}{0.5\columnwidth}
		\centering
		\includegraphics[width=\columnwidth]{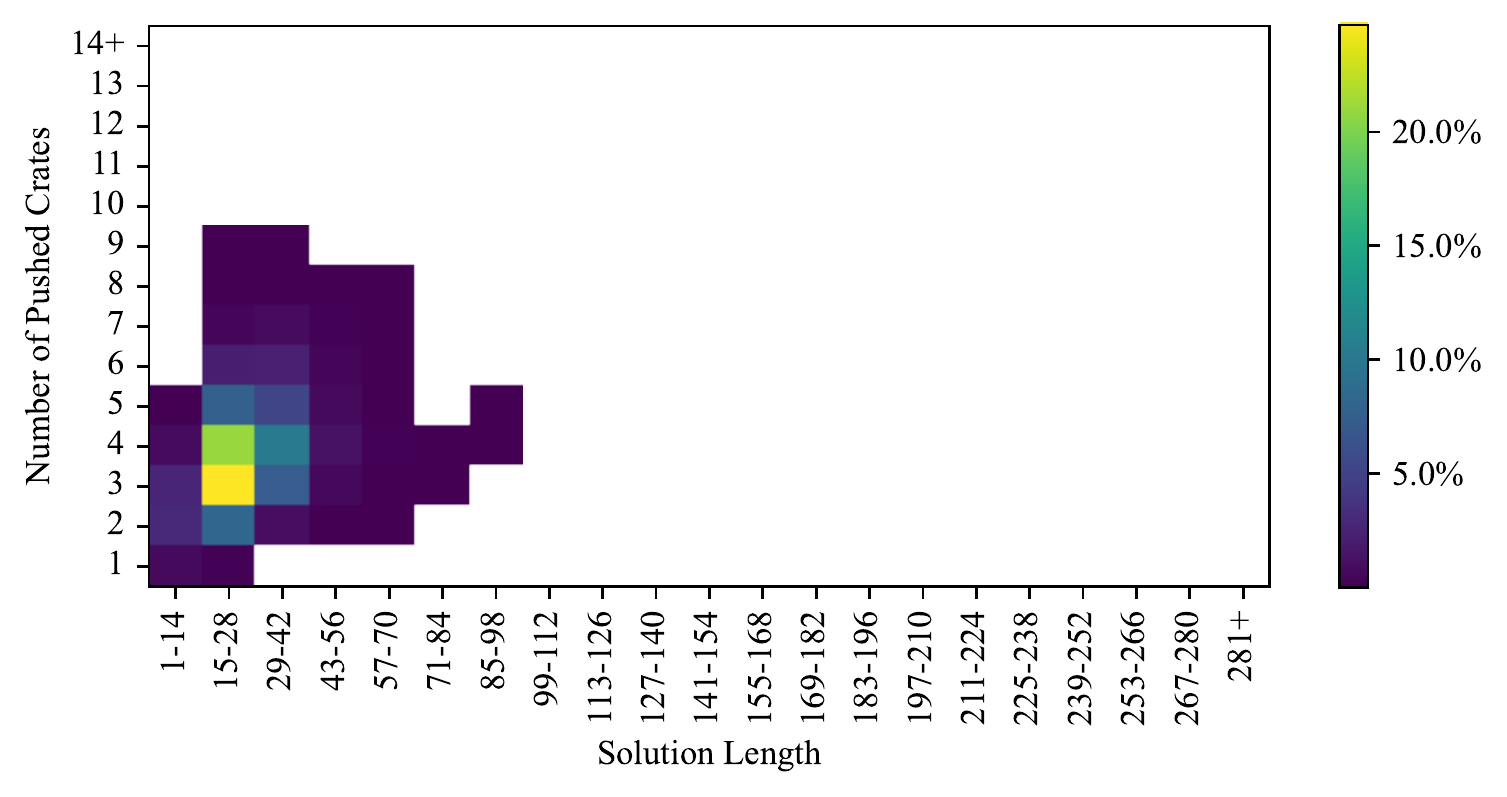}
		\caption{DS (Signatures)}
	\end{subfigure}
	
	\begin{subfigure}{0.5\columnwidth}
		\centering
		\includegraphics[width=\columnwidth]{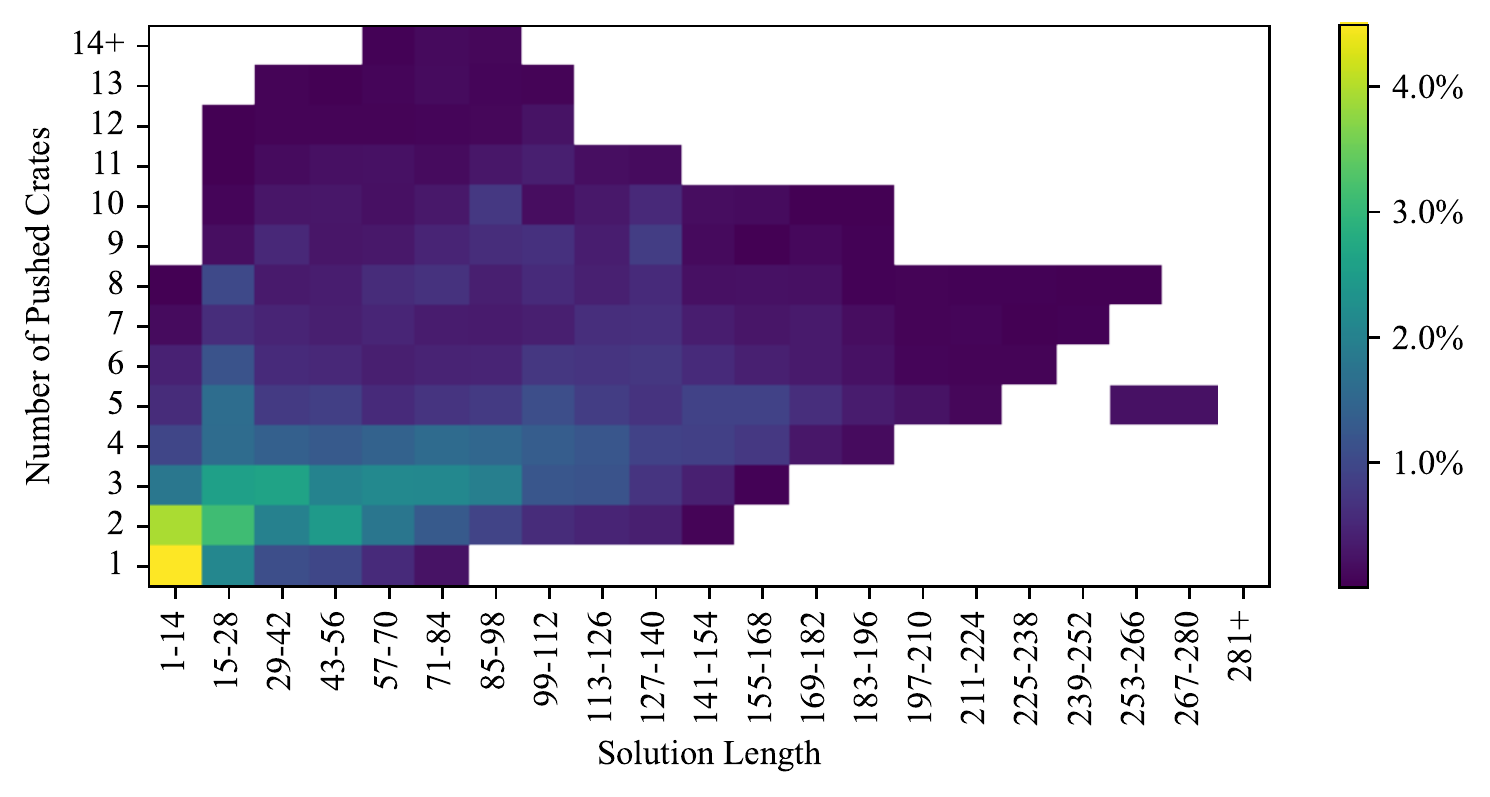}
		\caption{DS}
	\end{subfigure}~
	\begin{subfigure}{0.5\columnwidth}
		\centering
		\includegraphics[width=\columnwidth]{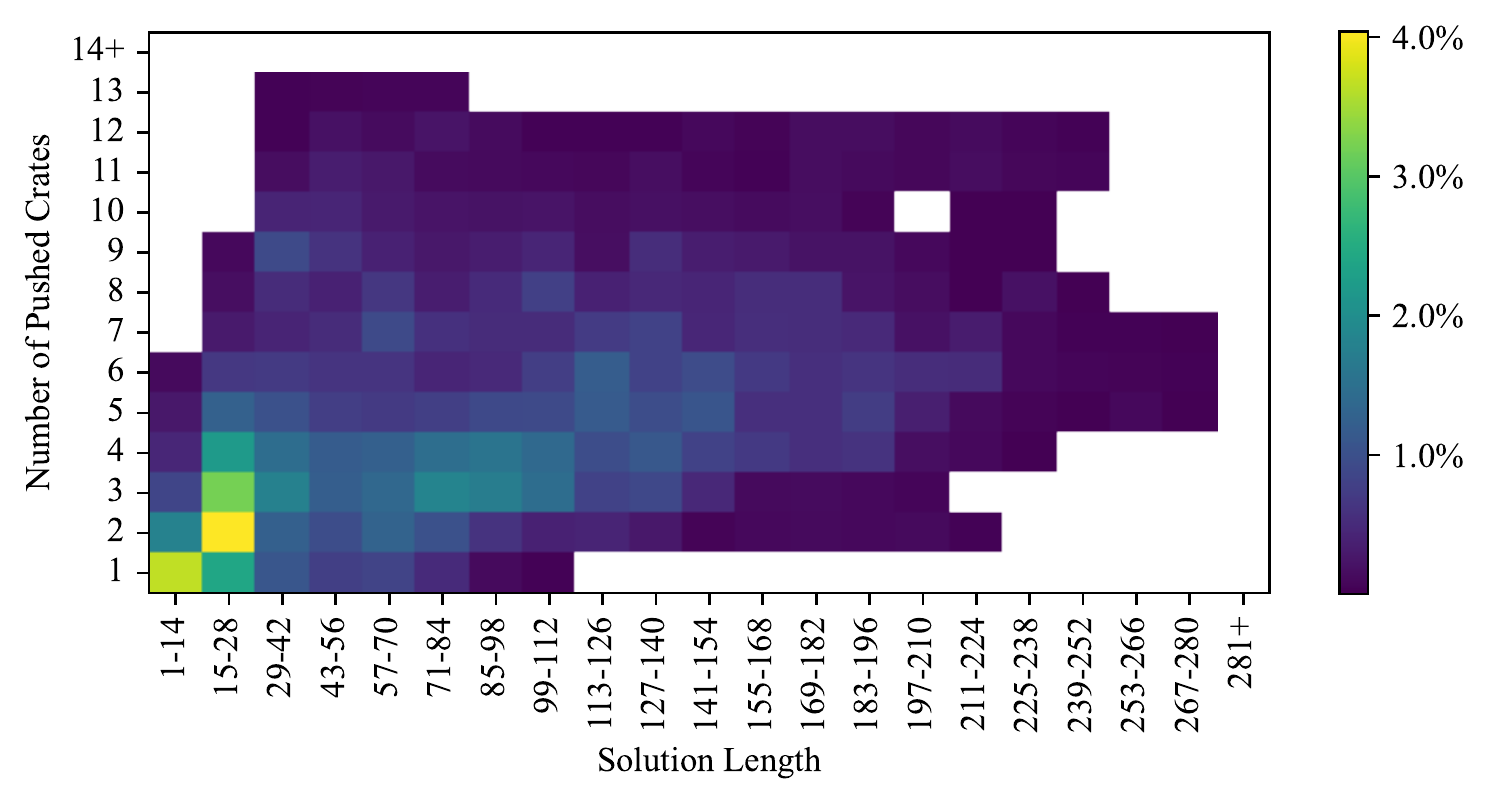}
		\caption{DS + PR}
	\end{subfigure}
	
	\begin{subfigure}{0.5\columnwidth}
		\centering
		\includegraphics[width=\columnwidth]{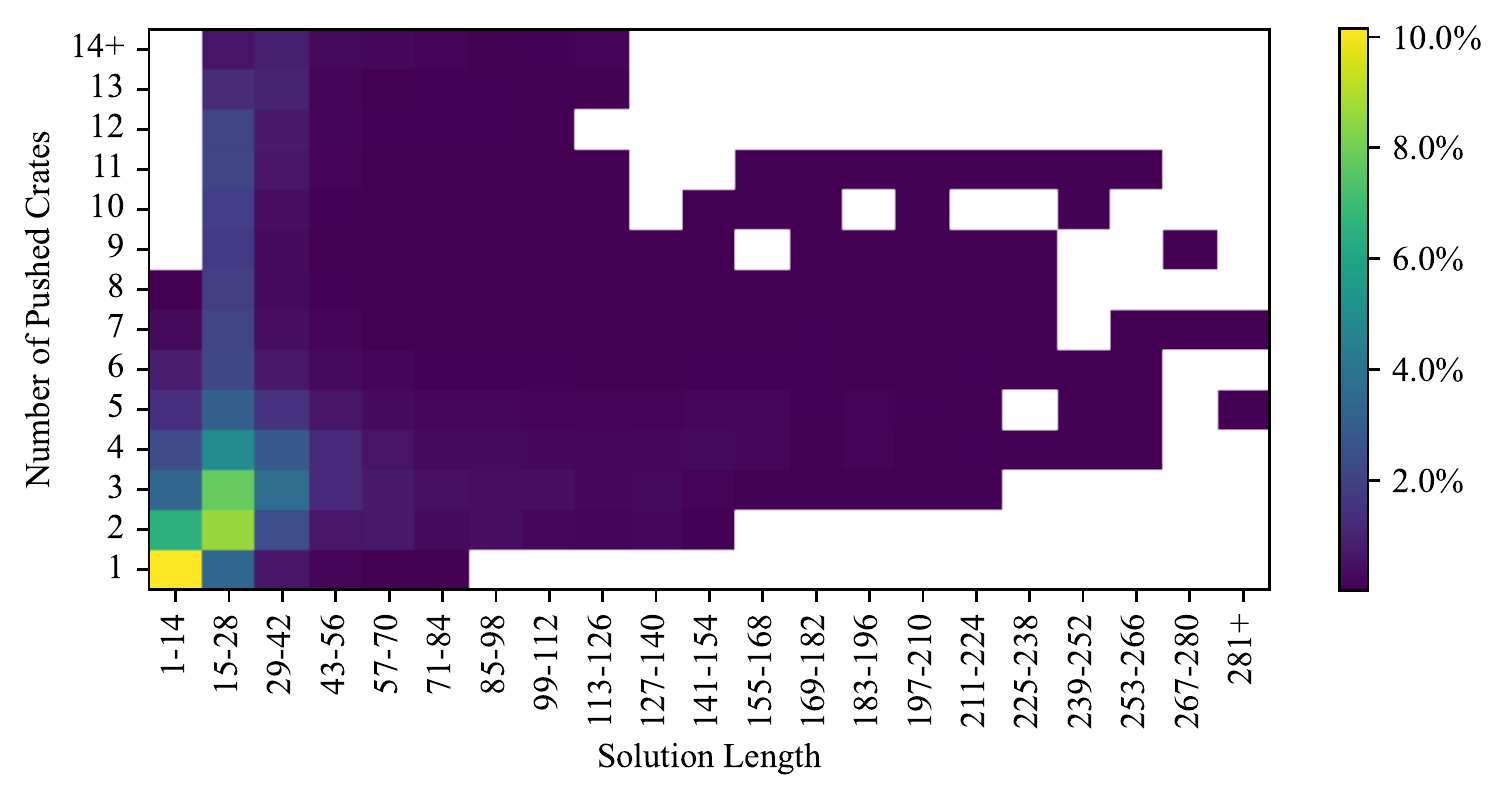}
		\caption{DS + AUG}
	\end{subfigure}~
	\begin{subfigure}{0.5\columnwidth}
		\centering
		\includegraphics[width=\columnwidth]{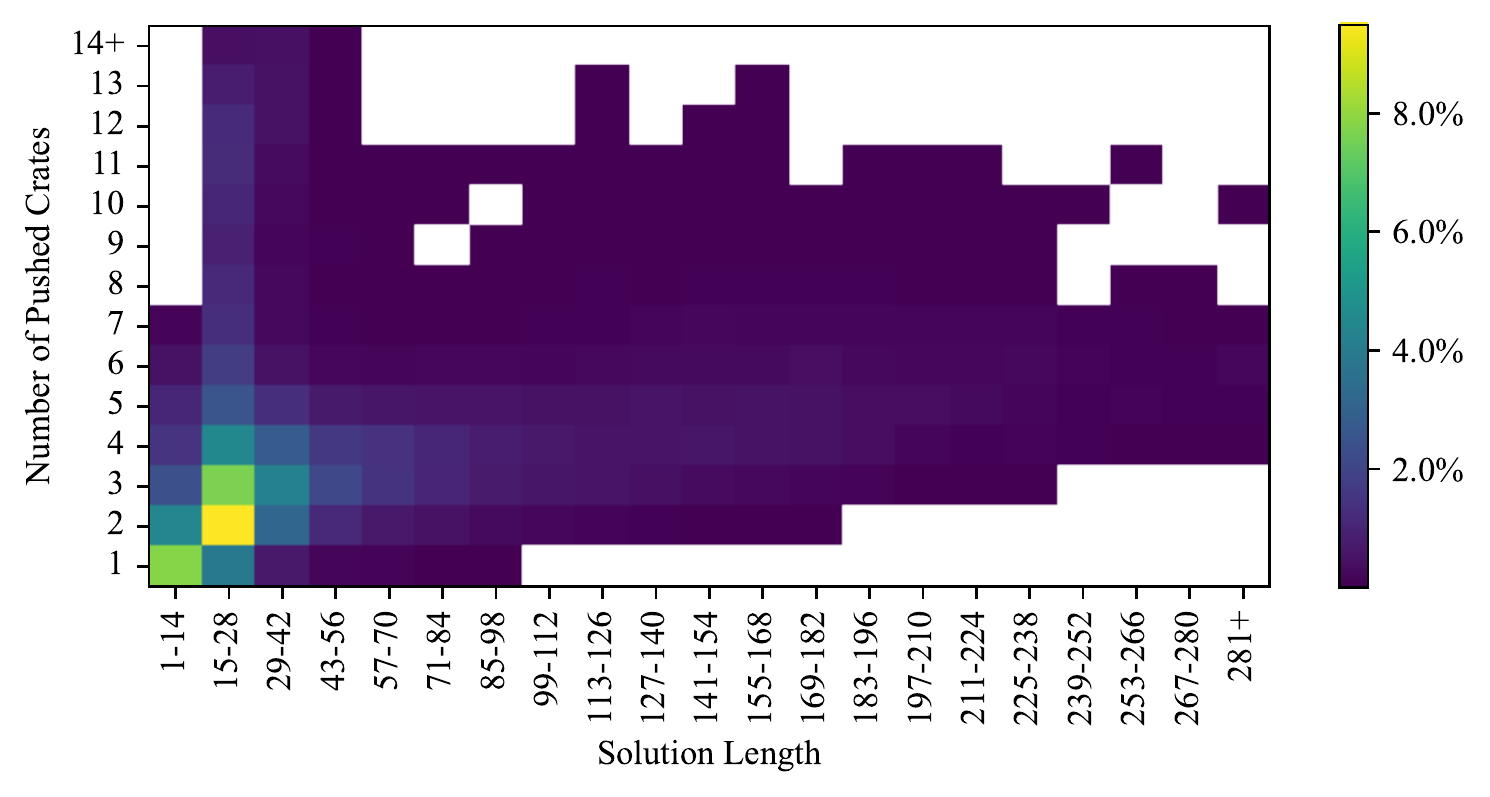}
		\caption{DS + PR + AUG}
	\end{subfigure}
	
	\caption{The Expressive Ranges of the Sokoban Generators at the size $7 \times 7$. DS: Diversity Sampling. PR: Property Reward. AUG: Data Augmentation.}
	\label{fig:sokoban_er}
\end{figure}

As shown in \tablename\ \ref{tbl:sokoban_qd}, the playability statistics are $>50\%$ for the in-training sizes $5 \times 5$, $6 \times 6$ \& $7 \times 7$. Training without diversity sampling yields the highest playability and high tile-diversity too. However, the solution diversity is low as shown by the unique signature and the solution length statistics. Diversity sampling significantly improves the solution diversity at the cost of a significant playability decrease. Although the signature key yields high unique signatures, the solution lengths only cover a small range. Using the tuple key significantly increases the solution length range despite decreasing the unique signatures. Adding a property reward further increases the solution length range. Data augmentation increases the tile diversity and unique signatures but decreases the playability and the average solution length. The duplicate percentage was omitted since it is always $<0.1\%$, except when using the tuple key without data augmentation, where the duplicates were nearly $1.3\%$. So, data augmentation also decreases the duplicates. \figurename\ \ref{fig:sokoban_er} shows the expressive ranges \cite{Smith2010ExRange} of the Sokoban generators where the x-axis is the solution length, and the y-axis is the number of pushed crates. The results show that the tuple key significantly expands the expressive range, and adding the property reward further expands the range along the solution length axis.

\begin{table}[t]
	\tiny
	\caption{Quality and Diversity of the Zelda Generators. DS: Diversity Sampling. PR: Property Reward. AUG: Data Augmentation. Out-of-training sizes are underlined.}
	\label{tbl:zelda_qd}
	\centering
	
\begin{tabular}{ @{\hspace{0.1cm}} c @{\hspace{0.1cm}} c @{\hspace{0.1cm}} r @{\hspace{0.2cm}} r @{\hspace{0.2cm}} r @{\hspace{0.2cm}} r @{\hspace{0.2cm}} r @{\hspace{0.2cm}} r @{\hspace{0.1cm}} }
	\toprule
	& AUG &&\multicolumn{1}{c}{$7\times 11$} & \multicolumn{1}{c}{$\underline{6\times 10}$} & \multicolumn{1}{c}{$\underline{10\times 6}$} & \multicolumn{1}{c}{$\underline{8\times 12}$} & \multicolumn{1}{c}{$\underline{9\times 13}$}\\
	\midrule
	\multirow{4}{*}{\begin{tabular}{c}Zelda \\ \textsc{(DS+PR)} \end{tabular}} & \multirow{2}{*}{.} & Playable\% & $49.2\% \pm 11.7\%$ & $\pmb{59.2\% \pm 7.2\%}$ & $52.4\% \pm 8.3\%$ & $60.5\% \pm 11.1\%$ & $60.5\% \pm 12.7\%$\\
	&  & Tile Diversity & $0.40 \pm 0.02$ & $0.44 \pm 0.02$ & $0.45 \pm 0.04$ & $0.40 \pm 0.02$ & $0.40 \pm 0.02$\\
	\cmidrule{2-8}
	& \multirow{2}{*}{\checkmark} & Playable\% & $\pmb{73.4\% \pm 4.0\%}$ & $57.8\% \pm 5.3\%$ & $\pmb{72.0\% \pm 7.2\%}$ & $\pmb{78.2\% \pm 4.4\%}$ & $\pmb{74.5\% \pm 6.3\%}$\\
	&  & Tile Diversity & $\pmb{0.44 \pm 0.01}$ & $\pmb{0.47 \pm 0.02}$ & $\pmb{0.46 \pm 0.01}$ & $\pmb{0.42 \pm 0.02}$ & $\pmb{0.41 \pm 0.03}$\\
	\bottomrule
\end{tabular}
	
\end{table}

\begin{table}[t]
	\tiny
	\caption{Quality and Diversity of the Danger Dave Generators. DS: Diversity Sampling. AUG: Data Augmentation. Out-of-training sizes are underlined.}
	\label{tbl:dave_qd}
	\centering
	
	\begin{tabular}{ @{\hspace{0.1cm}} c @{\hspace{0.1cm}} c @{\hspace{0.1cm}} r @{\hspace{0.2cm}} r @{\hspace{0.2cm}} r @{\hspace{0.2cm}} r @{\hspace{0.2cm}} r @{\hspace{0.2cm}} r @{\hspace{0.1cm}} }
		\toprule
		& AUG &&\multicolumn{1}{c}{$7\times 11$} & \multicolumn{1}{c}{$\underline{6\times 10}$} & \multicolumn{1}{c}{$\underline{10\times 6}$} & \multicolumn{1}{c}{$\underline{8\times 12}$} & \multicolumn{1}{c}{$\underline{9\times 13}$}\\
		\midrule
		\multirow{4}{*}{\begin{tabular}{c}Danger \\ Dave \\ (DS) \end{tabular}} & \multirow{2}{*}{.} & Playable\% & $51.4\% \pm 11.7\%$ & $\pmb{45.7\% \pm 15.3\%}$ & $27.5\% \pm 21.0\%$ & $\pmb{35.9\% \pm 20.3\%}$ & $\pmb{27.6\% \pm 22.1\%}$\\
		&  & Tile Diversity & $0.40 \pm 0.04$ & $0.44 \pm 0.07$ & $0.47 \pm 0.04$ & $0.38 \pm 0.08$ & $0.37 \pm 0.10$\\
		\cmidrule{2-8}
		& \multirow{2}{*}{\checkmark} & Playable\% & $\pmb{54.4\% \pm 8.6\%}$ & $31.4\% \pm 8.2\%$ & $\pmb{30.7\% \pm 12.3\%}$ & $30.5\% \pm 7.9\%$ & $21.2\% \pm 9.0\%$\\
		&  & Tile Diversity & $\pmb{0.48 \pm 0.03}$ & $\pmb{0.53 \pm 0.05}$ & $\pmb{0.52 \pm 0.03}$ & $\pmb{0.47 \pm 0.04}$ & $\pmb{0.45 \pm 0.06}$\\
		\bottomrule
	\end{tabular}
	
\end{table}

\begin{figure*}[p]
	\centering
	
	\begin{subfigure}{0.5\columnwidth}
		\centering
		\includegraphics[width=\columnwidth]{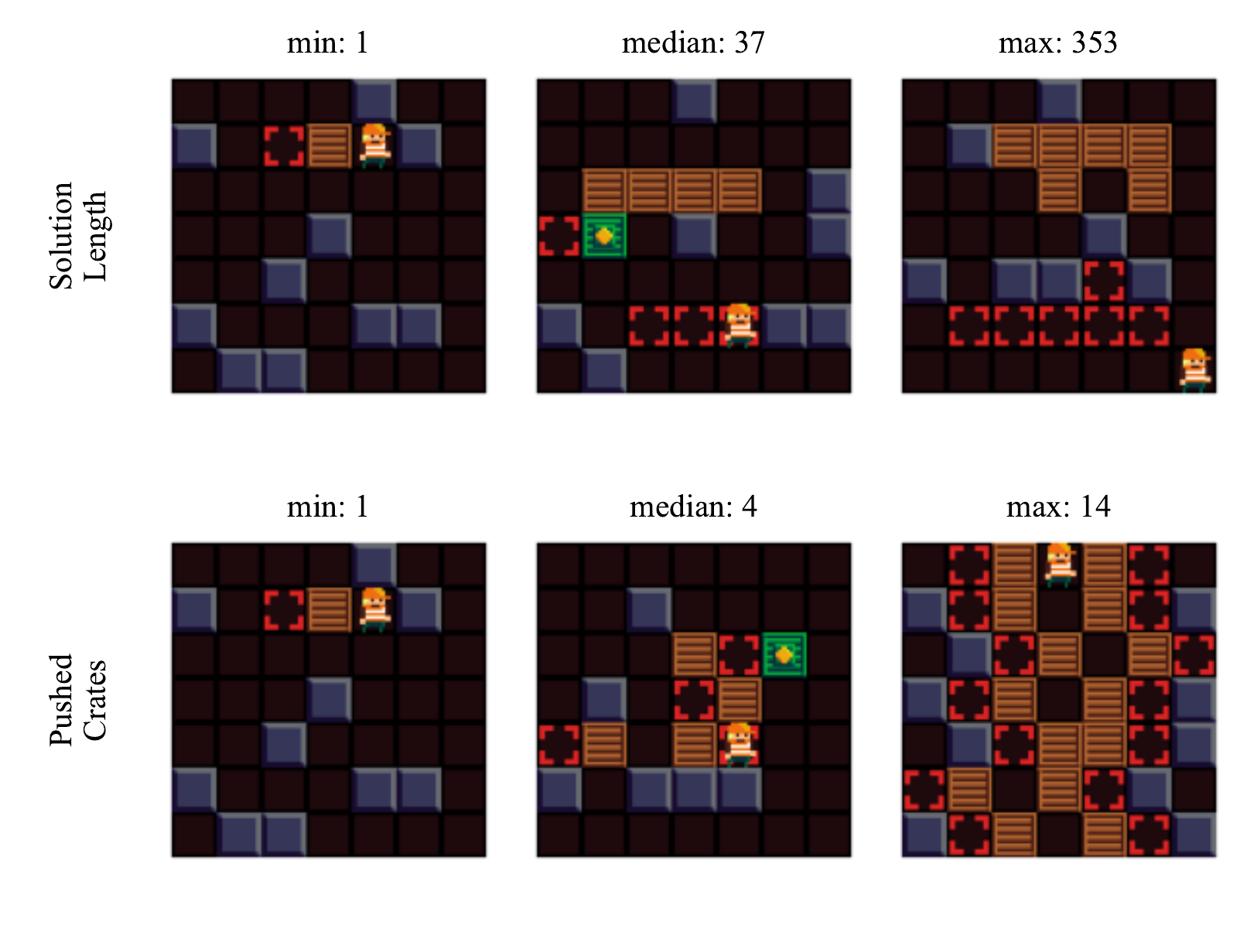}
		\caption{Sokoban $7 \times 7$ (DS + PR + AUG)}
	\end{subfigure}~
	\begin{subfigure}{0.5\columnwidth}
		\centering
		\includegraphics[width=\columnwidth]{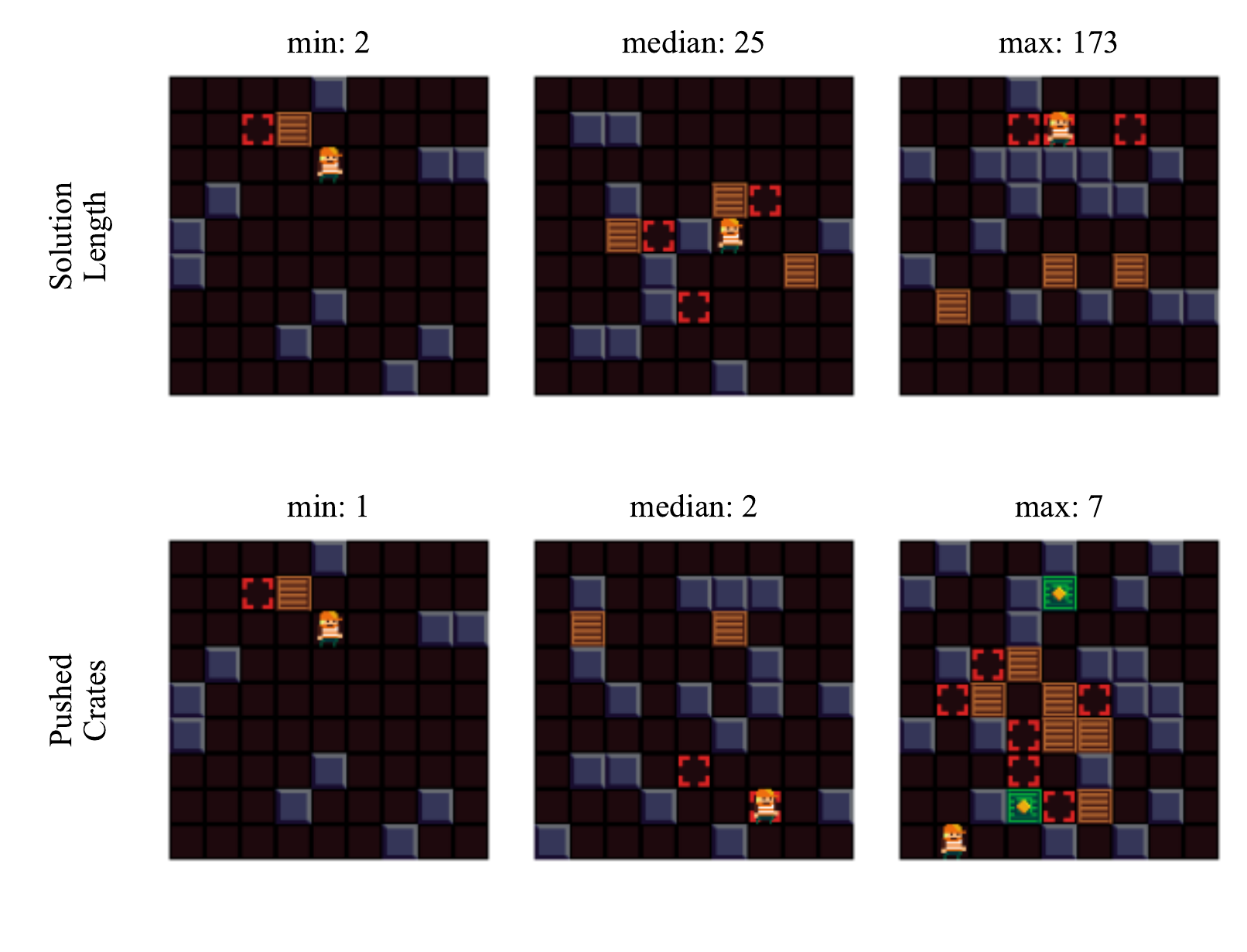}
		\caption{Sokoban $9 \times 9$ (DS + PR + AUG)}
	\end{subfigure}
	
	\begin{subfigure}{0.5\columnwidth}
		\centering
		\includegraphics[width=\columnwidth]{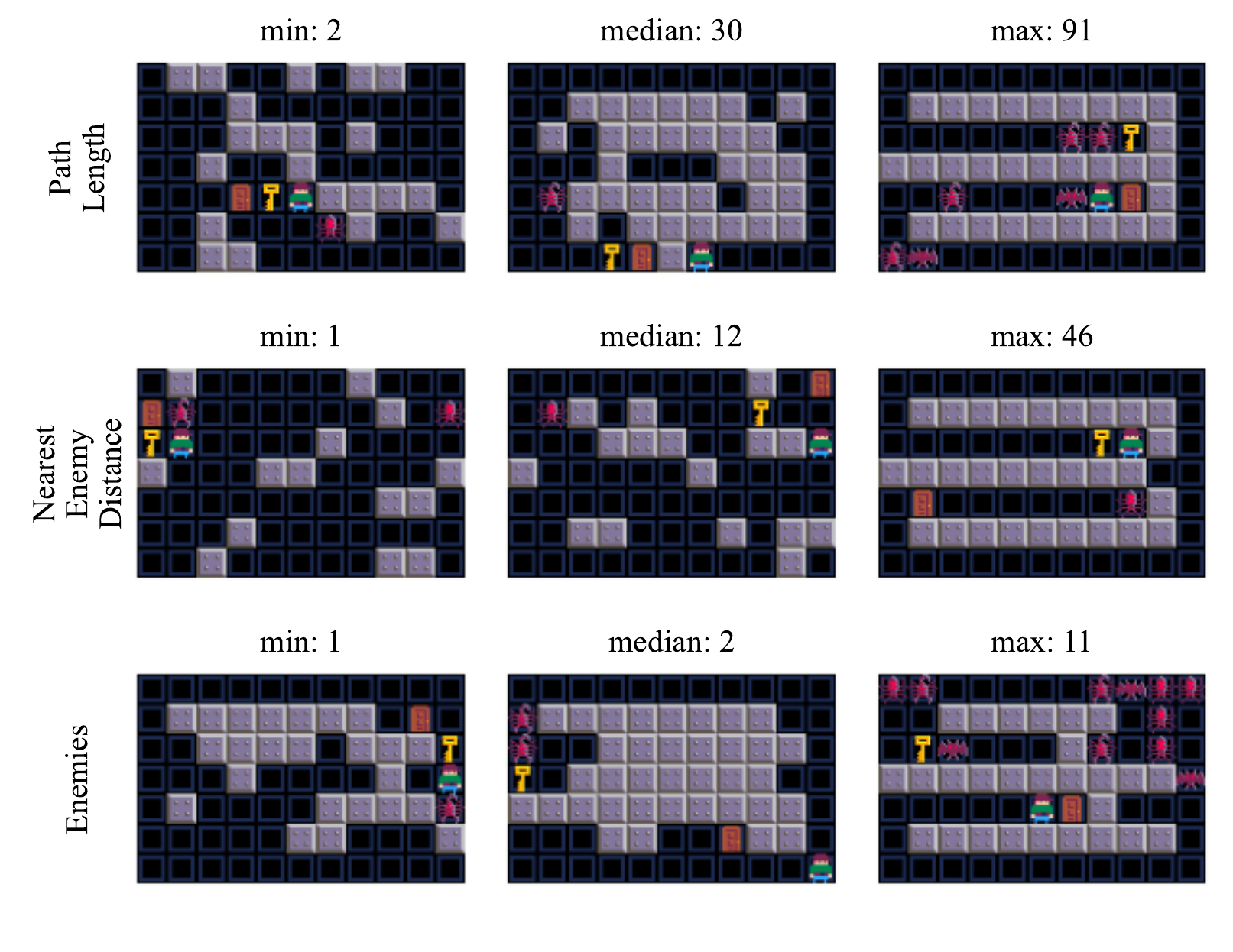}
		\caption{Zelda $7 \times 11$ (DS + PR + AUG)}
	\end{subfigure}~
	\begin{subfigure}{0.5\columnwidth}
		\centering
		\includegraphics[width=\columnwidth]{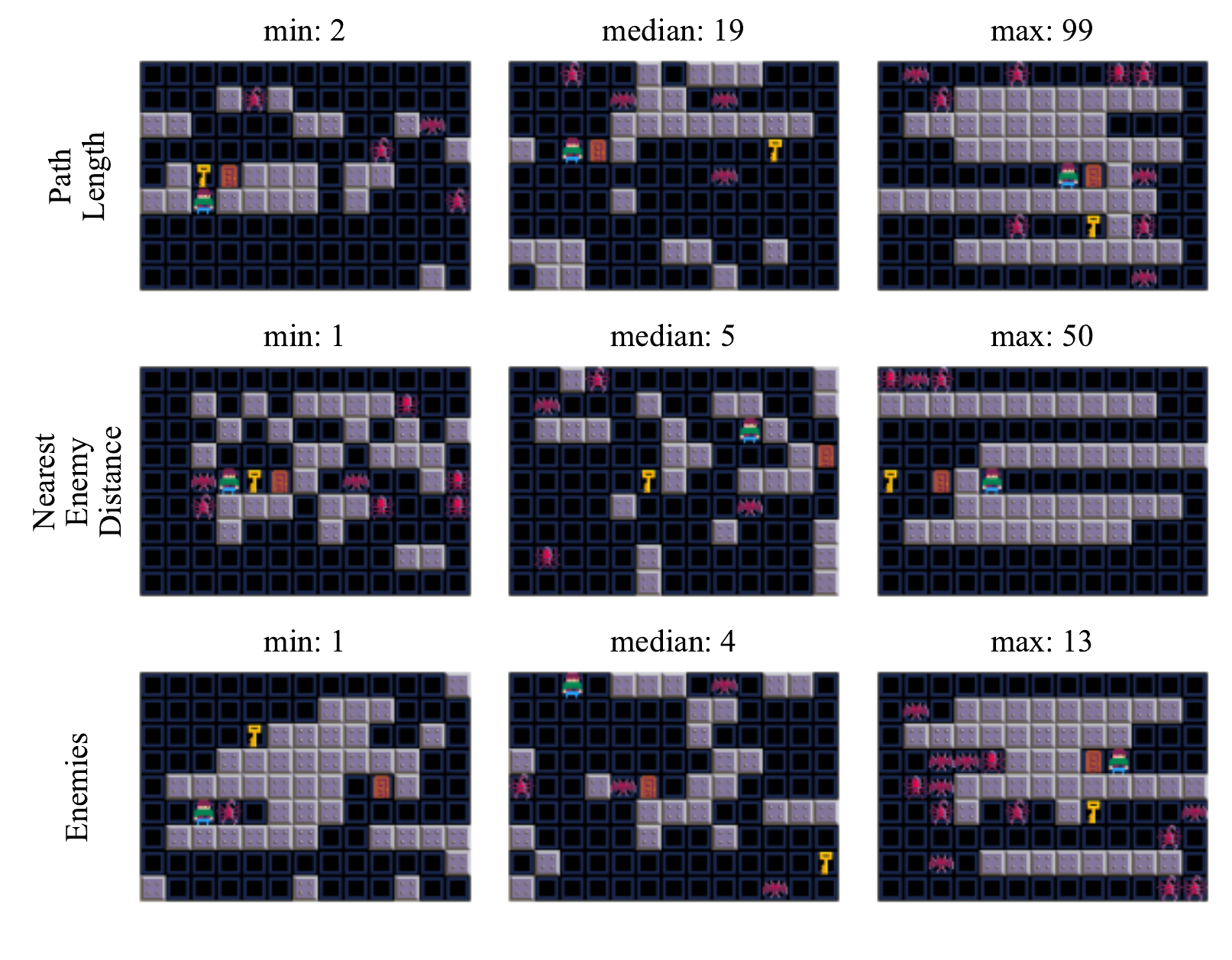}
		\caption{Zelda $9 \times 13$ (DS + PR + AUG)}
	\end{subfigure}
	
	\begin{subfigure}{0.5\columnwidth}
		\centering
		\includegraphics[width=\columnwidth]{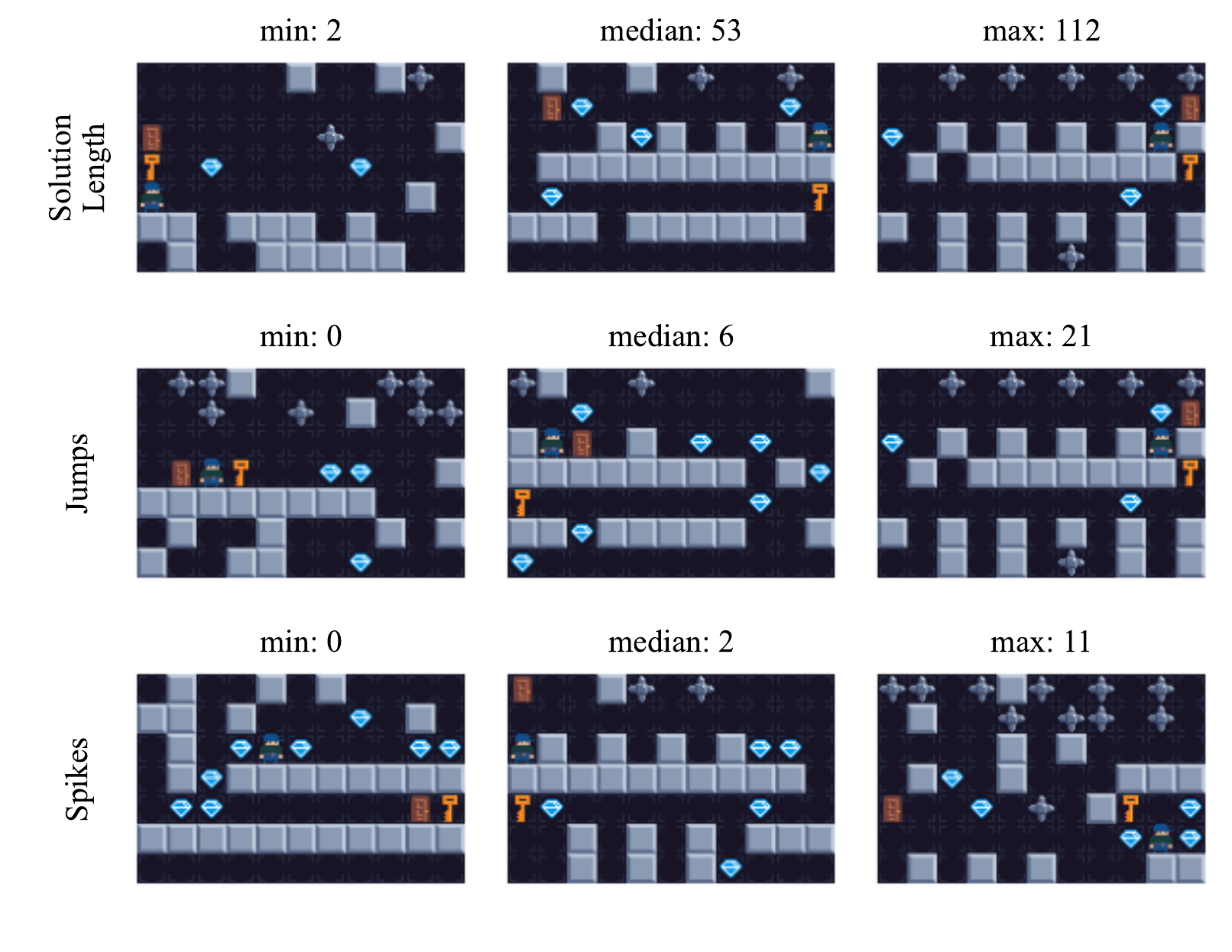}
		\caption{Danger Dave $7 \times 11$  (DS + AUG)}
	\end{subfigure}~
	\begin{subfigure}{0.5\columnwidth}
		\centering
		\includegraphics[width=\columnwidth]{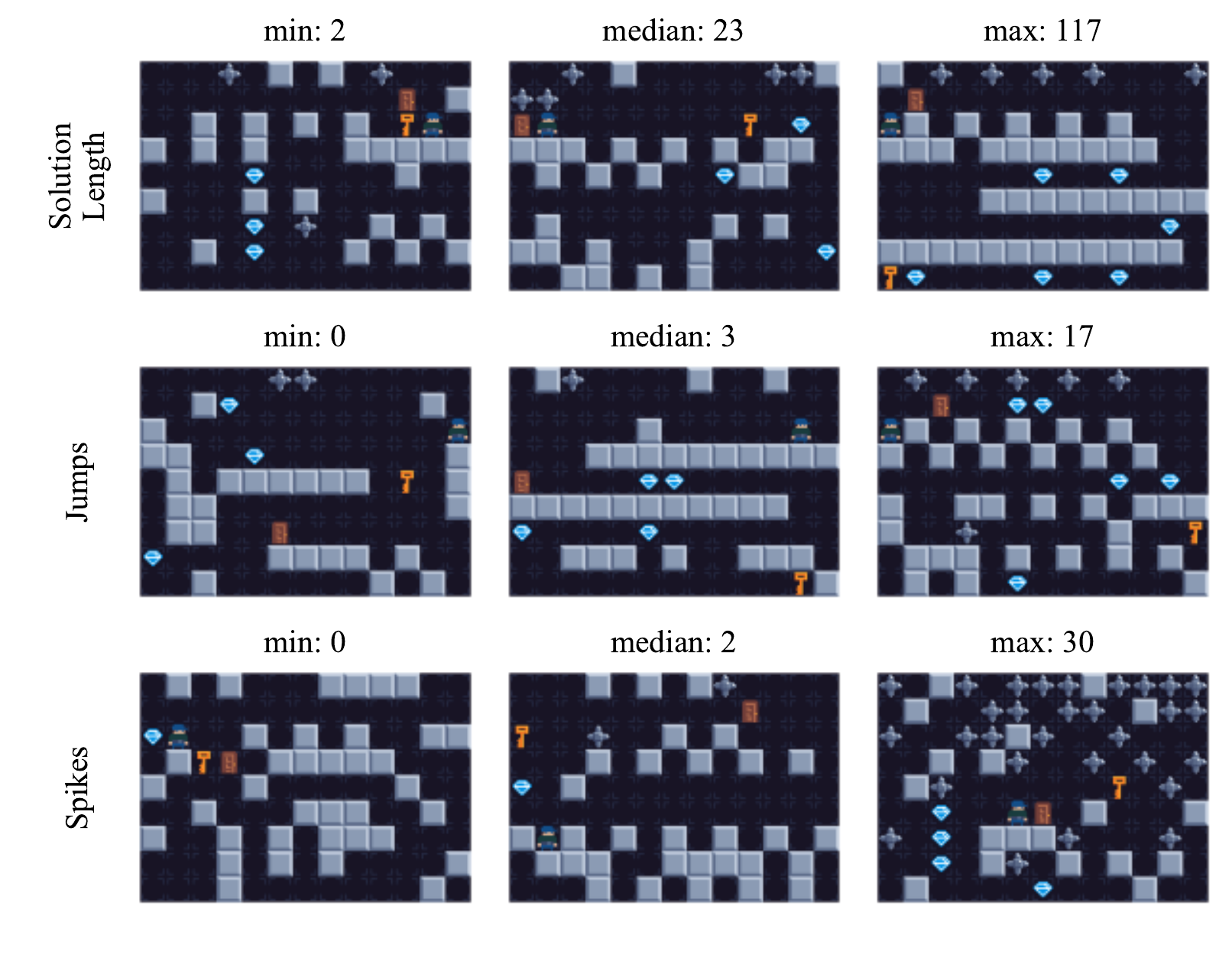}
		\caption{Danger Dave $9 \times 13$  (DS + AUG)}
	\end{subfigure}
	
	\caption{A Sample of the generated Levels. DS: Diversity Sampling. PR: Property Reward. AUG: Data Augmentation.}
	\label{fig:level_samples}
\end{figure*}

%\begin{figure*}[t]
%	\centering
%	
%	\begin{subfigure}{0.5\columnwidth}
%		\centering
%		\includegraphics[width=\columnwidth]{figures/samples/SOKOBAN/GFLOW}
%		\caption{No DS}
%	\end{subfigure}~
%	\begin{subfigure}{0.5\columnwidth}
%		\centering
%		\includegraphics[width=\columnwidth]{figures/samples/SOKOBAN/DIVPUSHSIG_GFLOW}
%		\caption{DS (Signatures)}
%	\end{subfigure}
%	
%%	\begin{subfigure}{0.5\columnwidth}
%%		\centering
%%		\includegraphics[width=\columnwidth]{figures/samples/SOKOBAN/DIV_GFLOW}
%%		\caption{DS}
%%	\end{subfigure}~
%	\begin{subfigure}{0.5\columnwidth}
%		\centering
%		\includegraphics[width=\columnwidth]{figures/samples/SOKOBAN/PREW_DIV_GFLOW}
%		\caption{DS + PR}
%	\end{subfigure}~
%%	\begin{subfigure}{0.5\columnwidth}
%%		\centering
%%		\includegraphics[width=\columnwidth]{figures/samples/SOKOBAN/AUG_DIV_GFLOW}
%%		\caption{DS + AUG}
%%	\end{subfigure}~
%	\begin{subfigure}{0.5\columnwidth}
%		\centering
%		\includegraphics[width=\columnwidth]{figures/samples/SOKOBAN/AUG_PREW_DIV_GFLOW}
%		\caption{DS + PR + AUG}
%	\end{subfigure}
%	
%	\caption{A Sample of $7 \times 7$ Generated Sokoban Levels. DS, PR \& AUG stands for Diversity Sampling, Property Reward and Data Augmentation respectively.}
%	\label{fig:sokoban_samples}
%\end{figure*}

\tablename\ \ref{tbl:zelda_qd} and \tablename\ \ref{tbl:dave_qd} show the results for Zelda and Danger Dave respectively (both using Diversity Sampling), where sizes smaller than $7 \times 11$ were omitted to leave space for the out-of-training sizes. The results support the claim that data augmentation increases the tile diversity. It also shows that data augmentation does not necessarily decrease the playability as observed in the results for Zelda.

\figurename\ \ref{fig:level_samples} shows generated level samples where the levels for each game are sampled from the same model. To show the extent of the generator's range, the sample contains the minimum, median and maximum along the controlled properties. The figure contains an out-of-training size for each game to demonstrate the generators' diversity for out-of-training sizes.

\tablename\ \ref{tbl:sokoban_qd}, \tablename\ \ref{tbl:zelda_qd}, \tablename\ \ref{tbl:dave_qd} and \figurename\ \ref{fig:level_samples} show that our method can generate levels for out-of-training sizes, however the performance can decline, as observed in the results for Sokoban and Danger Dave. In Sokoban, the decline is more prominent when the tuple key is used, which could suggest that the generator was unable to generalize the knowledge it gained for generating levels with more crates and longer solutions to out-of-training sizes.

%\begin{figure*}[t]
%	\centering
%	
%	\begin{subfigure}{0.5\columnwidth}
%		\centering
%		\includegraphics[width=\columnwidth]{figures/ranges/SOKOBAN/DIVPUSHSIG_GFLOW/div_7x7}
%		\caption{DS (Signatures)}
%	\end{subfigure}~
%	\begin{subfigure}{0.5\columnwidth}
%		\centering
%		\includegraphics[width=\columnwidth]{figures/ranges/SOKOBAN/PREW_DIV_GFLOW/div_7x7}
%		\caption{DS + PR}
%	\end{subfigure}
%	
%	\caption{Tile Diversity of the Sokoban Generators at the size $7 \times 7$. DS \& PR stands for Diversity Sampling and Property Reward respectively.}
%	\label{fig:sokoban_div}
%\end{figure*}

\subsection{Controllability}\label{ssec:controllability}

\begin{table}[t]
	\tiny
	\caption{Control Test Results of the Sokoban Generators. DS: Diversity Sampling. PR: Property Reward. AUG: Data Augmentation. Out-of-training sizes are underlined.}
	\label{tbl:sokoban_cont}
	\centering

\begin{tabular}{c @{\hspace{0.1cm}} c @{\hspace{0.1cm}} c @{\hspace{0.1cm}} r @{\hspace{0.1cm}} r @{\hspace{0.2cm}} r @{\hspace{0.2cm}} r @{\hspace{0.2cm}} r @{\hspace{0.2cm}} r @{\hspace{0.2cm}} r}
	\toprule
	& & & & \multicolumn{3}{c}{Pushed Crates}&\multicolumn{3}{c}{Solution Length}\\
	\cmidrule(lr){5-7}\cmidrule(lr){8-10}
	DS & PR & AUG & &\multicolumn{1}{c}{$7\times 7$} & \multicolumn{1}{c}{$\underline{8\times 8}$} & \multicolumn{1}{c}{$\underline{9\times 9}$} & \multicolumn{1}{c}{$7\times 7$} & \multicolumn{1}{c}{$\underline{8\times 8}$} & \multicolumn{1}{c}{$\underline{9\times 9}$}\\
	\midrule
	\multirow{4}{*}{\checkmark }&\multirow{4}{*}{ . }&\multirow{4}{*}{ .} & Playable\% & $\pmb{64.6\% \pm 4.2\%}$ & $27.3\% \pm 17.1\%$ & $18.2\% \pm 11.0\%$ & $\pmb{62.9\% \pm 5.6\%}$ & $35.5\% \pm 9.5\%$ & $24.1\% \pm 7.8\%$\\
	& &  & Avg. Error & $0.51 \pm 0.09$ & $1.44 \pm 0.66$ & $2.41 \pm 0.74$ & $11.92 \pm 1.43$ & $21.03 \pm 6.42$ & $26.24 \pm 5.87$\\
	& &  & $R^2$ & $0.91 \pm 0.04$ & $0.22 \pm 0.41$ & $-0.59 \pm 0.56$ & $0.68 \pm 0.09$ & $-0.03 \pm 0.54$ & $-0.53 \pm 0.58$\\
	& &  & Score & $49.0\% \pm 4.5\%$ & $13.3\% \pm 9.2\%$ & $5.6\% \pm 3.5\%$ & $\pmb{20.4\% \pm 4.0\%}$ & $\pmb{7.7\% \pm 2.9\%}$ & $3.8\% \pm 1.8\%$\\
	\midrule
	\multirow{4}{*}{\checkmark }&\multirow{4}{*}{ \checkmark }&\multirow{4}{*}{ .} & Playable\% & $64.0\% \pm 7.1\%$ & $27.5\% \pm 13.8\%$ & $\pmb{20.6\% \pm 12.1\%}$ & $62.3\% \pm 4.7\%$ & $\pmb{36.2\% \pm 6.2\%}$ & $25.7\% \pm 7.2\%$\\
	& &  & Avg. Error & $\pmb{0.45 \pm 0.18}$ & $1.47 \pm 0.98$ & $2.15 \pm 1.01$ & $\pmb{11.72 \pm 1.41}$ & $23.58 \pm 8.04$ & $28.88 \pm 8.28$\\
	& &  & $R^2$ & $\pmb{0.92 \pm 0.04}$ & $0.14 \pm 0.63$ & $-0.51 \pm 0.72$ & $\pmb{0.69 \pm 0.09}$ & $-0.27 \pm 0.69$ & $-0.77 \pm 0.72$\\
	& &  & Score & $\pmb{50.1\% \pm 5.7\%}$ & $11.8\% \pm 2.8\%$ & $6.2\% \pm 3.5\%$ & $19.4\% \pm 3.1\%$ & $6.8\% \pm 1.7\%$ & $3.9\% \pm 1.6\%$\\
	\midrule
	\multirow{4}{*}{\checkmark }&\multirow{4}{*}{ . }&\multirow{4}{*}{ \checkmark} & Playable\% & $48.1\% \pm 3.2\%$ & $25.7\% \pm 6.2\%$ & $16.8\% \pm 7.5\%$ & $46.6\% \pm 8.2\%$ & $34.0\% \pm 7.7\%$ & $\pmb{28.4\% \pm 8.7\%}$\\
	& &  & Avg. Error & $0.67 \pm 0.12$ & $1.49 \pm 0.44$ & $2.26 \pm 0.96$ & $18.43 \pm 4.82$ & $24.03 \pm 6.09$ & $26.76 \pm 3.14$\\
	& &  & $R^2$ & $0.88 \pm 0.03$ & $0.41 \pm 0.42$ & $-0.53 \pm 1.33$ & $0.24 \pm 0.38$ & $-0.26 \pm 0.37$ & $-0.45 \pm 0.24$\\
	& &  & Score & $32.6\% \pm 2.4\%$ & $11.7\% \pm 3.2\%$ & $4.8\% \pm 3.4\%$ & $11.4\% \pm 1.5\%$ & $6.9\% \pm 2.2\%$ & $\pmb{4.9\% \pm 1.9\%}$\\
	\midrule
	\multirow{4}{*}{\checkmark }&\multirow{4}{*}{ \checkmark }&\multirow{4}{*}{ \checkmark} & Playable\% & $47.1\% \pm 6.3\%$ & $\pmb{28.8\% \pm 3.8\%}$ & $17.9\% \pm 5.5\%$ & $45.8\% \pm 5.3\%$ & $34.9\% \pm 6.3\%$ & $22.6\% \pm 6.8\%$\\
	& &  & Avg. Error & $0.55 \pm 0.27$ & $\pmb{0.87 \pm 0.46}$ & $\pmb{1.42 \pm 0.43}$ & $14.87 \pm 2.25$ & $\pmb{19.87 \pm 3.23}$ & $\pmb{22.05 \pm 4.47}$\\
	& &  & $R^2$ & $0.90 \pm 0.08$ & $\pmb{0.73 \pm 0.25}$ & $\pmb{0.20 \pm 0.30}$ & $0.53 \pm 0.11$ & $\pmb{0.10 \pm 0.25}$ & $\pmb{-0.15 \pm 0.30}$\\
	& &  & Score & $35.3\% \pm 8.4\%$ & $\pmb{18.4\% \pm 4.4\%}$ & $\pmb{7.6\% \pm 3.4\%}$ & $12.3\% \pm 2.9\%$ & $7.4\% \pm 1.7\%$ & $4.3\% \pm 1.3\%$\\
	\bottomrule
\end{tabular}
	
\end{table}

\tablename\ \ref{tbl:sokoban_cont}, \tablename\ \ref{tbl:zelda_cont} and \tablename\ \ref{tbl:dave_cont} present the control statistics of the Sokoban, Zelda and Danger Dave generators, respectively. The tables contain the following metrics:
\begin{itemize}
	\item `Playable\%' presents the playability which is the percentage of generated levels that satisfy the functional requirements.
	\item `Avg. Error' is the mean absolute error between the requested and the generated levels' properties which is computed for the playable levels only.
	\item `$R^2$' is the coefficient of determination where the request controls are treated as the true values and the generated level properties are treated as the predicted values. It was added because the average error is hard to compare without considering the variance in the control values.
	\item `Score' is the control score \cite{Zakaria2022plgdl} where the tolerance is 2 for the pushed crate count and 10 for the solution length. This is reported for Sokoban to facilitate the comparison of our results with the ones in \cite{Zakaria2022plgdl}. We did not report the control score for the other games to save space.
\end{itemize}
In \tablename\ \ref{tbl:sokoban_cont}, for the in-training size $7 \times 7$, data augmentation decreases the playability and increase the average error, thus decreasing the score. The effect is different for the out-of-training sizes, where data augmentation usually improves the results. Regardless, the performance still declines when the requested size is out-of-training. The decline can cause the $R^2$ values to be close to (or less than) zero. In addition, the standard deviations of all the metrics are usually higher for out-of-training sizes, which means that the results tends to highly vary across runs.

\begin{table}[t]
	\tiny
	\caption{Control Test Results of the Zelda Generators. DS: Diversity Sampling. PR: Property Reward. AUG: Data Augmentation. Out-of-training sizes are underlined.}
	\label{tbl:zelda_cont}
	\centering

\begin{tabular}{c @{\hspace{0.1cm}} c @{\hspace{0.1cm}} r @{\hspace{0.1cm}} r @{\hspace{0.2cm}} r @{\hspace{0.2cm}} r @{\hspace{0.2cm}} r @{\hspace{0.2cm}} r}
	\toprule
	& Control & &\multicolumn{1}{c}{$7\times 11$} & \multicolumn{1}{c}{$\underline{6\times 10}$} & \multicolumn{1}{c}{$\underline{10\times 6}$} & \multicolumn{1}{c}{$\underline{8\times 12}$} & \multicolumn{1}{c}{$\underline{9\times 13}$}\\
	\midrule
	\multirow{9}{*}{\rotatebox[origin=c]{90}{\begin{tabular}{ @{\hspace{0.1cm}} c @{\hspace{0.1cm}} }Zelda \\ (DS + PR + AUG) \end{tabular}}} & \multirow{3}{*}{Enemies} & Playable\% & $70.4\% \pm 4.1\%$ & $56.4\% \pm 5.3\%$ & $69.6\% \pm 7.2\%$ & $71.5\% \pm 7.6\%$ & $69.4\% \pm 8.5\%$\\
	&  & Avg. Error & $0.52 \pm 0.39$ & $0.52 \pm 0.21$ & $0.55 \pm 0.24$ & $0.83 \pm 0.54$ & $1.18 \pm 0.70$\\
	&  & $R^2$ & $0.92 \pm 0.08$ & $0.91 \pm 0.04$ & $0.90 \pm 0.06$ & $0.86 \pm 0.14$ & $0.77 \pm 0.24$\\
	\cmidrule(lr){2-8}
	& \multirow{3}{*}{\begin{tabular}{c}Nearest Enemy\\ Distance\end{tabular}} & Playable\% & $65.4\% \pm 6.0\%$ & $40.0\% \pm 9.2\%$ & $61.3\% \pm 10.6\%$ & $54.9\% \pm 7.0\%$ & $43.4\% \pm 10.3\%$\\
	&  & Avg. Error & $4.62 \pm 0.30$ & $3.72 \pm 0.35$ & $5.30 \pm 0.81$ & $6.75 \pm 0.80$ & $8.82 \pm 1.48$\\
	&  & $R^2$ & $0.51 \pm 0.09$ & $0.32 \pm 0.13$ & $0.10 \pm 0.30$ & $0.19 \pm 0.15$ & $-0.10 \pm 0.15$\\
	\cmidrule(lr){2-8}
	& \multirow{3}{*}{Path Length} & Playable\% & $68.2\% \pm 3.7\%$ & $43.6\% \pm 10.9\%$ & $62.0\% \pm 12.6\%$ & $61.9\% \pm 5.2\%$ & $49.1\% \pm 9.4\%$\\
	&  & Avg. Error & $8.81 \pm 0.57$ & $7.41 \pm 0.71$ & $9.55 \pm 1.36$ & $13.80 \pm 1.77$ & $17.40 \pm 2.46$\\
	&  & $R^2$ & $0.60 \pm 0.07$ & $0.39 \pm 0.10$ & $0.25 \pm 0.20$ & $0.25 \pm 0.24$ & $0.01 \pm 0.22$\\
	\bottomrule
\end{tabular}

\end{table}

\begin{table}[t]
	\tiny
	\caption{Control Test Results of the Danger Dave Generators. DS: Diversity Sampling. AUG: Data Augmentation. Out-of-training sizes are underlined.}
	\label{tbl:dave_cont}
	\centering

	\begin{tabular}{c @{\hspace{0.1cm}} c @{\hspace{0.1cm}} r @{\hspace{0.1cm}} r @{\hspace{0.2cm}} r @{\hspace{0.2cm}} r @{\hspace{0.2cm}} r @{\hspace{0.2cm}} r}
		\toprule
		& Control & &\multicolumn{1}{c}{$7\times 11$} & \multicolumn{1}{c}{$\underline{6\times 10}$} & \multicolumn{1}{c}{$\underline{10\times 6}$} & \multicolumn{1}{c}{$\underline{8\times 12}$} & \multicolumn{1}{c}{$\underline{9\times 13}$}\\
		\midrule
		\multirow{9}{*}{\rotatebox[origin=c]{90}{\begin{tabular}{ @{\hspace{0.1cm}} c @{\hspace{0.1cm}} }Danger \\ Dave (DS + AUG) \end{tabular}}} & \multirow{3}{*}{Spikes} & Playable\% & $37.2\% \pm 9.2\%$ & $22.1\% \pm 10.5\%$ & $21.9\% \pm 9.9\%$ & $19.2\% \pm 9.6\%$ & $15.3\% \pm 7.4\%$\\
		&  & Avg. Error & $0.50 \pm 0.16$ & $0.43 \pm 0.15$ & $0.68 \pm 0.31$ & $0.76 \pm 0.25$ & $1.16 \pm 0.41$\\
		&  & $R^2$ & $0.89 \pm 0.04$ & $0.89 \pm 0.07$ & $0.84 \pm 0.05$ & $0.83 \pm 0.10$ & $0.78 \pm 0.13$\\
		\cmidrule(lr){2-8}
		& \multirow{3}{*}{Jumps} & Playable\% & $55.8\% \pm 10.5\%$ & $21.4\% \pm 4.8\%$ & $23.1\% \pm 11.5\%$ & $18.9\% \pm 5.9\%$ & $11.4\% \pm 4.8\%$\\
		&  & Avg. Error & $1.53 \pm 0.21$ & $2.11 \pm 0.62$ & $2.44 \pm 0.47$ & $3.23 \pm 0.90$ & $5.17 \pm 1.49$\\
		&  & $R^2$ & $0.83 \pm 0.05$ & $0.38 \pm 0.28$ & $0.25 \pm 0.29$ & $0.25 \pm 0.33$ & $-0.08 \pm 0.30$\\
		\cmidrule(lr){2-8}
		& \multirow{3}{*}{Solution Length} & Playable\% & $52.7\% \pm 7.1\%$ & $28.0\% \pm 9.4\%$ & $28.8\% \pm 12.5\%$ & $28.1\% \pm 10.2\%$ & $17.1\% \pm 10.2\%$\\
		&  & Avg. Error & $7.85 \pm 1.05$ & $7.47 \pm 1.79$ & $7.20 \pm 1.40$ & $13.23 \pm 2.82$ & $19.42 \pm 5.01$\\
		&  & $R^2$ & $0.75 \pm 0.07$ & $0.56 \pm 0.22$ & $0.65 \pm 0.14$ & $0.48 \pm 0.25$ & $0.13 \pm 0.44$\\
		\bottomrule
	\end{tabular}
	
\end{table}

In \tablename\ \ref{tbl:zelda_cont} and \tablename\ \ref{tbl:dave_cont}, only the results with data augmentation were reported to save space, since they were usually better especially at out-of-training sizes. The results show some performance decline at out-of-training sizes, but it is not as significant as seen in the results for Sokoban, which could be attributed to the use of tailored GMMs. Some out-of-training sizes have better results compared to the reported in-training size. In addition, the average $R^2$ values are rarely near or below zero. An interesting observation is that the best control results are for tile-count oriented controls (Pushed crates in Sokoban, Enemies in Zelda and Spikes in Danger Dave). This can be attributed to the recurrent models' ability to keep count of items in its memory.

\subsection{Training and Generation Time}\label{ssec:time}

\begin{table}[t]
	\tiny
	\caption{Training and Generation Times of the Sokoban Generators. DS: Diversity Sampling. PR: Property Reward. AUG: Data Augmentation. `Sig.' is an abbreviation for the solution signature. Out-of-training sizes are underlined.}
	\label{tbl:sokoban_time}
	\centering
	
	\begin{tabular}{@{\hspace{0.1cm}} c @{\hspace{0.1cm}} c @{\hspace{0.1cm}} c @{\hspace{0.4cm}} c @{\hspace{0.4cm}} r @{\hspace{0.2cm}} r @{\hspace{0.2cm}} r @{\hspace{0.2cm}} r @{\hspace{0.2cm}} r @{\hspace{0.1cm}} }
		\toprule
		& & & & \multicolumn{5}{c}{Generation Time (Batch Size = 100 levels)}\\
		\cmidrule(lr){5-9}
		DS & PR & AUG & Training Time &\multicolumn{1}{c}{$5\times 5$} & \multicolumn{1}{c}{$6\times 6$} & \multicolumn{1}{c}{$7\times 7$} & \multicolumn{1}{c}{$\underline{8\times 8}$} & \multicolumn{1}{c}{$\underline{9\times 9}$}\\
		\midrule
		.  &  .  &  . & $3h\ 20min\ 13s$ $\pm $ $59min\ 18s$ & $0.35 s \pm 0.03 s$ & $0.74 s \pm 0.15 s$ & $0.83 s \pm 0.19 s$ & $0.88 s \pm 0.37 s$ & $0.89 s \pm 0.51 s$\\
		Sig.  &  .  &  . & $7h\ 14min\ 25s$ $\pm $ $32min\ 16s$ & $0.66 s \pm 0.08 s$ & $1.76 s \pm 0.15 s$ & $2.27 s \pm 0.23 s$ & $2.80 s \pm 0.19 s$ & $2.65 s \pm 0.78 s$\\
		\checkmark  &  .  &  . & $5h\ 54min\ 45s$ $\pm $ $22min\ 50s$ & $0.47 s \pm 0.08 s$ & $1.49 s \pm 0.11 s$ & $1.91 s \pm 0.14 s$ & $1.58 s \pm 0.30 s$ & $1.52 s \pm 0.27 s$\\
		\checkmark  &  \checkmark  &  . & $5h\ 48min\ 54s$ $\pm $ $31min\ 52s$ & $0.43 s \pm 0.06 s$ & $1.46 s \pm 0.19 s$ & $1.82 s \pm 0.18 s$ & $1.92 s \pm 0.84 s$ & $1.90 s \pm 0.87 s$\\
		\checkmark  &  .  &  \checkmark & $5h\ 39min\ 56s$ $\pm $ $12min\ 28s$ & $0.48 s \pm 0.03 s$ & $1.24 s \pm 0.07 s$ & $1.60 s \pm 0.11 s$ & $2.01 s \pm 0.29 s$ & $2.13 s \pm 0.60 s$\\
		\checkmark  &  \checkmark  &  \checkmark & $6h\ 14min\ 25s$ $\pm $ $38min\ 13s$ & $0.52 s \pm 0.06 s$ & $1.32 s \pm 0.10 s$ & $1.70 s \pm 0.15 s$ & $1.89 s \pm 0.19 s$ & $1.93 s \pm 0.32 s$\\
		\bottomrule
	\end{tabular}
	
\end{table}

All the experiments use the same model architecture, except for the input size, which depends on the number of controls, and the output size which equals the tileset size. However, the effects of the input and output sizes were unnoticeable in our measurements. During generation, the model call is divided as follows: $1$ conditional-embedding module call, followed by $wh$ calls to the recurrent cells and the action module (where $wh$ is the level area). So, we measured the module call time as a function of the level area by timing 1000 generation requests (batch size = 1) at various sizes for every game, and applying linear regression. The generation time turned out to be $0.81 wh + 0.55$ ms on average (the correlation coefficient $r = 0.9999$).

In practice, level generation can be batched to generate multiple levels in parallel. Also, since the generated level is not guaranteed to be playable, it should also be verified by an automated solver. The computational cost to verify a level will depend on the game and the search space size of the generated levels. Therefore, we generate 5 batches (batch size = 100) from each generator at a variety of sizes, and measure the average time required to generate and verify the levels. The results are reported in \tablename\ \ref{tbl:sokoban_time}, \tablename\ \ref{tbl:zelda_time} and \tablename\ \ref{tbl:dave_time} for Sokoban, Zelda and Danger Dave respectively. As expected, the generation time increases as the level area increases. Additionally, Sokoban requires the longest time, since its levels' search spaces are larger than those of Zelda and Danger Dave. Among the Sokoban generators, those trained without diversity sampling require the least time, since most of the generated levels are trivial to solve, while those trained with diversity sampling using the signature key requires the longest time, since most of the generated levels pass the basic functional requirements (e.g., having only one player) and require invoking the search procedure.

\begin{table}[t]
	\tiny
	\caption{Training and Generation Times of the Zelda Generators (with Diversity Sampling and Property Reward). AUG: Data Augmentation. Out-of-training sizes are underlined.}
	\label{tbl:zelda_time}
	\centering
	
	\begin{tabular}{ @{\hspace{0.1cm}} c @{\hspace{0.4cm}} c @{\hspace{0.4cm}} r @{\hspace{0.2cm}} r @{\hspace{0.2cm}} r @{\hspace{0.2cm}} r @{\hspace{0.2cm}} r @{\hspace{0.2cm}} }
		\toprule
		& & \multicolumn{5}{c}{Generation Time (Batch Size = 100 levels)}\\
		\cmidrule(lr){3-7}
		AUG & Training Time &\multicolumn{1}{c}{$7\times 11$} & \multicolumn{1}{c}{$\underline{6\times 10}$} & \multicolumn{1}{c}{$\underline{10\times 6}$} & \multicolumn{1}{c}{$\underline{8\times 12}$} & \multicolumn{1}{c}{$\underline{9\times 13}$}\\
		\midrule
		. & $3h\ 29min\ 06s$ $\pm $ $08min\ 27s$ & $0.10 s \pm 0.02 s$ & $0.08 s \pm 0.01 s$ & $0.08 s \pm 0.01 s$ & $0.12 s \pm 0.01 s$ & $0.15 s \pm 0.01 s$\\
		\checkmark & $3h\ 33min\ 47s$ $\pm $ $10min\ 14s$ & $0.10 s \pm 0.01 s$ & $0.08 s \pm 0.01 s$ & $0.08 s \pm 0.01 s$ & $0.13 s \pm 0.01 s$ & $0.16 s \pm 0.01 s$\\
		\bottomrule
	\end{tabular}
	
\end{table}

\begin{table}[t]
	\tiny
	\caption{Training and Generation Times of the Danger Dave Generators (with Diversity Sampling). AUG: Data Augmentation. Out-of-training sizes are underlined.}
	\label{tbl:dave_time}
	\centering
	
	\begin{tabular}{ @{\hspace{0.1cm}} c @{\hspace{0.4cm}} c @{\hspace{0.4cm}} r @{\hspace{0.2cm}} r @{\hspace{0.2cm}} r @{\hspace{0.2cm}} r @{\hspace{0.2cm}} r @{\hspace{0.2cm}} }
		\toprule
		& & \multicolumn{5}{c}{Generation Time (Batch Size = 100 levels)}\\
		\cmidrule(lr){3-7}
		AUG & Training Time &\multicolumn{1}{c}{$7\times 11$} & \multicolumn{1}{c}{$\underline{6\times 10}$} & \multicolumn{1}{c}{$\underline{10\times 6}$} & \multicolumn{1}{c}{$\underline{8\times 12}$} & \multicolumn{1}{c}{$\underline{9\times 13}$}\\
		\midrule
		. & $3h\ 25min\ 21s$ $\pm $ $04min\ 52s$ & $0.12 s \pm 0.03 s$ & $0.10 s \pm 0.01 s$ & $0.07 s \pm 0.01 s$ & $0.13 s \pm 0.02 s$ & $0.15 s \pm 0.02 s$\\
		\checkmark & $3h\ 34min\ 27s$ $\pm $ $02min\ 51s$ & $0.14 s \pm 0.01 s$ & $0.09 s \pm 0.01 s$ & $0.09 s \pm 0.01 s$ & $0.16 s \pm 0.02 s$ & $0.17 s \pm 0.03 s$\\
		\bottomrule
	\end{tabular}
	
\end{table}

All the experiments run for the same number of iterations, yet the training time differs due to other factors: the number \& the area of the training sizes, and the computational cost to verify the generated levels. Therefore, \tablename\ \ref{tbl:sokoban_time} shows the same trend observed in Sokoban's generation times, where training without diversity sampling requires the least training time, and training with diversity sampling using the signature key requires the longest training time. \tablename\ \ref{tbl:zelda_time} and \tablename\ \ref{tbl:dave_time} shows that training for Zelda and Danger Dave requires less time than training for Sokoban, despite having more sizes and larger level areas, because their functional requirements are faster to verify.

\subsection{Comparison with Controllable PCGRL}\label{ssec:comparison}

\begin{table}[t]
	\tiny
	\caption{Comparison between our generators and Controllable PCGRL for generating Sokoban levels of size $7\times 7$. $p$-values are computed between our 10-trials results and the Controllable PCGRL results using the Mann-Whitney U significance test.}
	\label{tbl:comparison}
	\centering
	
	\begin{tabular}{ c @{\hspace{0.1cm}} r @{\hspace{0.4cm}} r @{\hspace{0.2cm}} r @{\hspace{0.2cm}} r @{\hspace{0.2cm}} r @{\hspace{0.2cm}} }
		\toprule
		& & \multicolumn{2}{c}{Multi-Size GFlowNet (Ours)} &  &  \\
		\cmidrule(lr){3-4}
		 & & \multicolumn{1}{c}{1 Trial} &\multicolumn{1}{c}{10 Trials} & \multicolumn{1}{c}{C-PCGRL \cite{Earle2021ConPCGRL}} & \multicolumn{1}{c}{$p$-value \%} \\
		\midrule
		\multirow{3}{*}{\begin{tabular}{c}Quality \&\\ Diversity\end{tabular}} & Playable\% & $52.7\% \pm 3.6\%$ & $\pmb{99.9\% \pm 0.1\%}$ & $75.3\% \pm 7.7\%$ & $\pmb{1.19\%}$ \\
		& Diversity & $0.56 \pm 0.09$ & $0.56 \pm 0.09$ & $0.50 \pm 0.03$ & $30.95\%$ \\
		& Unique Sig. & $54.3\% \pm 6.0\%$ & $48.2\% \pm 6.3\%$ & $64.8\% \pm 13.9\%$ & $22.22\%$ \\
		\midrule
		\multirow{4}{*}{\begin{tabular}{c}Pushed\\ Crates\\ Control\end{tabular}}& Playable\% & $50.1\% \pm 4.5\%$ & $\pmb{99.2\% \pm 1.0\%}$ & $77.5\% \pm 8.0\%$ & $\pmb{0.79\%}$ \\
		& Avg. Error & $0.44 \pm 0.15$ & $\pmb{0.06 \pm 0.04}$ & $2.68 \pm 0.15$ & $\pmb{0.79\%}$ \\
		& $R^2$ & $0.93 \pm 0.03$ & $\pmb{0.99 \pm 0.01}$ & $-0.42 \pm 0.18$ & $\pmb{0.79\%}$ \\
		& Control Score & $39.16\% \pm 3.4\%$ & $\pmb{96.2\% \pm 2.9\%}$ & $19.73\% \pm 2.0\%$ & $\pmb{0.79\%}$ \\
		\midrule
		\multirow{4}{*}{\begin{tabular}{c}Solution\\ Length\\ Control\end{tabular}}& Playable\% & $47.9\% \pm 2.2\%$ & $\pmb{99.2\% \pm 0.4\%}$ & $75.6\% \pm 6.7\%$ & $\pmb{0.79\%}$ \\
		& Avg. Error & $14.49 \pm 2.30$ & $\pmb{7.35 \pm 1.43}$ & $18.99 \pm 5.62$ & $\pmb{0.79\%}$ \\
		& $R^2$ & $0.51 \pm 0.14$ & $\pmb{0.83 \pm 0.06}$ & $0.16 \pm 0.40$ & $\pmb{0.79\%}$ \\
		& Control Score & $14.6\% \pm 1.9\%$ & $\pmb{52.7\% \pm 4.4\%}$ & $19.98\% \pm 6.0\%$ & $\pmb{0.79\%}$ \\
		\midrule
		\multirow{3}{*}{\begin{tabular}{c}Training \&\\ Generation\\ Times\end{tabular}}& Training Time & $6h\ 43min\ \pm 2min$ & $\pmb{6h\ 43min\ \pm 2min}$ & $2\ days\ 20h\ \pm 7h$ & $\pmb{0.79\%}$ \\
		& Unc. Gen. Time & $0.76s\ \pm 0.05s$ & $\pmb{2.49s\ \pm 0.23s}$ & $73.63s\ \pm 5.82s$ & $\pmb{0.79\%}$ \\
		& Con. Gen. Time & $0.76s\ \pm 0.05s$ & $\pmb{7.62s\ \pm 0.50s}$ & $73.63s\ \pm 5.82s$ & $\pmb{0.79\%}$ \\
		\bottomrule
	\end{tabular}

\end{table}

In \tablename\ \ref{tbl:comparison}, we compare our method with Controllable PCGRL (C-PCGRL) \cite{Earle2021ConPCGRL} for Sokoban level generation, since it provides the most similar set of features by being controllable and able to learn without a dataset. Sokoban was picked over Zelda and Danger Dave for this comparison, since it is a more challenging benchmark as shown by the results presented in the previous subsections, and as discussed in \cite{Zakaria2022plgdl}. It should be noted that all the experiments we report in \tablename\ \ref{tbl:comparison} are run on the same machine which is different from the one used for the experiments reported in the previous subsections. Additionally, our generators included in this comparison are trained for $20,000$ iterations with diversity sampling (tuple key), property reward and data augmentation.

We present two sets of results for our generators. In the first set (denoted by 1 Trial in \tablename\ \ref{tbl:comparison}), the generator is given only one chance to generate a level so the results are measured as specified in section \ref{sec:expsetup}. However, as noted in \tablename\ \ref{tbl:comparison}, our generators, when limited to 1 trial, are $96.88\times$ faster at level generation compared to C-PCGRL. So, in the second set of results (denoted by  10 Trials in \tablename\ \ref{tbl:comparison}), we allow our generators to retry if their previous trial yielded an unplayable level (or does not satisfy the controls during the control tests) for up to 10 trials. Similar to Section \ref{ssec:time}, we report the generation and verification time for a batch of 100 levels, but we differentiate between uncontrollable generation (Unc. Gen.), where the generator stops as soon as it yields a playable level, and controllable generation (Con. Gen.) where the generator uses all the 10 trials to look for a level that is a better match for the requested controls (our code was not optimized to stop if the control error reaches 0). Therefore, the uncontrollable and controllable generation times are the same given 1 trial, but differ under 10 trials, where the controllable generation time increases 10 folds, while the uncontrollable generation time increases 3.28 folds only. In the last column of \tablename\ \ref{tbl:comparison}, we present the $p$-values calculated by the Mann-Whitney U test between our generators (10 Trials) and the C-PCGRL generators. Any results that are statistically significant ($p\text{-value} < 5\%$) are emphasized in the table.

\begin{figure*}[!t]
	\centering
	
	\begin{subfigure}{0.5\columnwidth}
		\centering
		\includegraphics[width=\columnwidth]{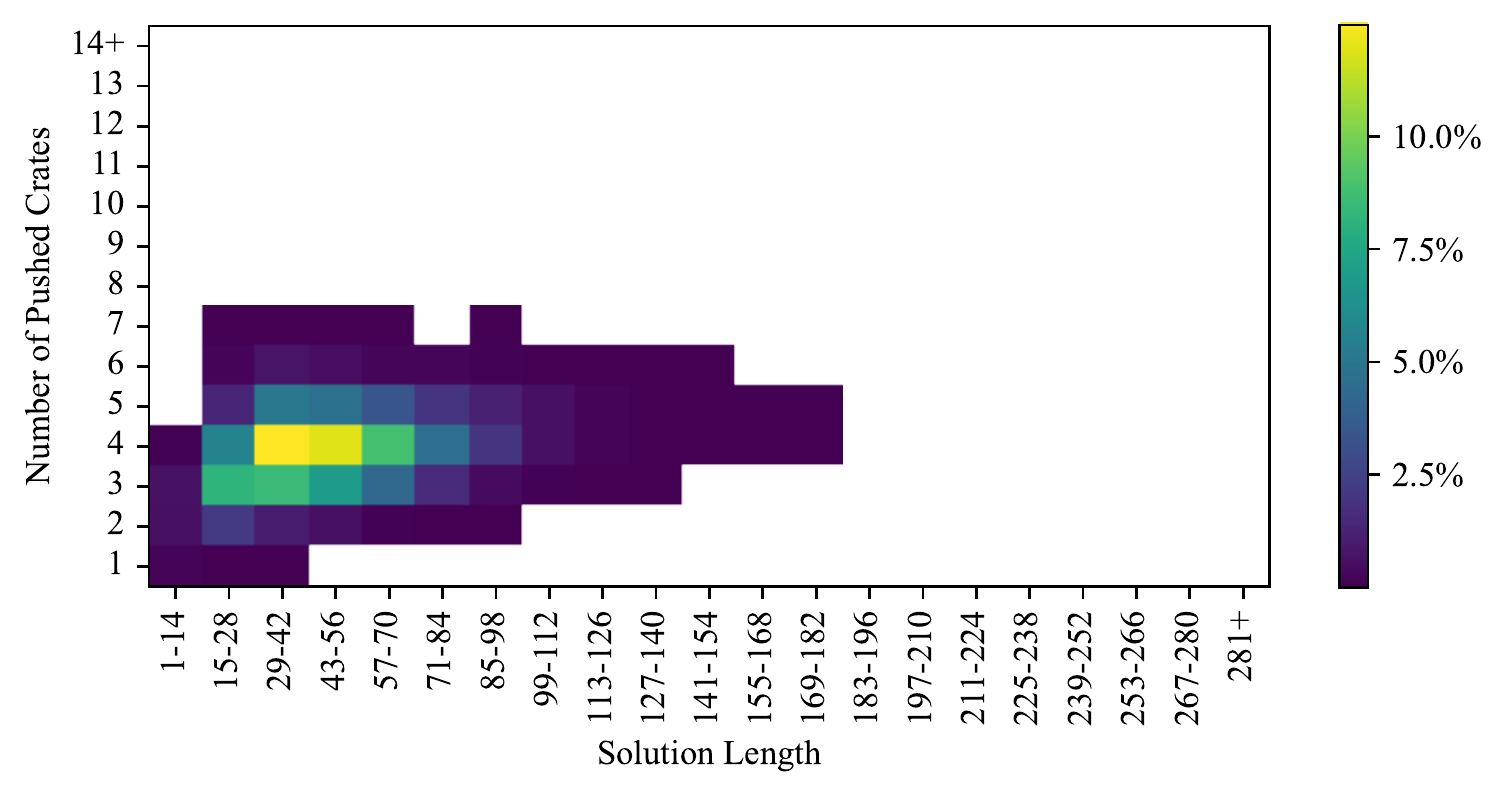}
		\caption{C-PCGRL}
	\end{subfigure}~
	\begin{subfigure}{0.5\columnwidth}
		\centering
		\includegraphics[width=\columnwidth]{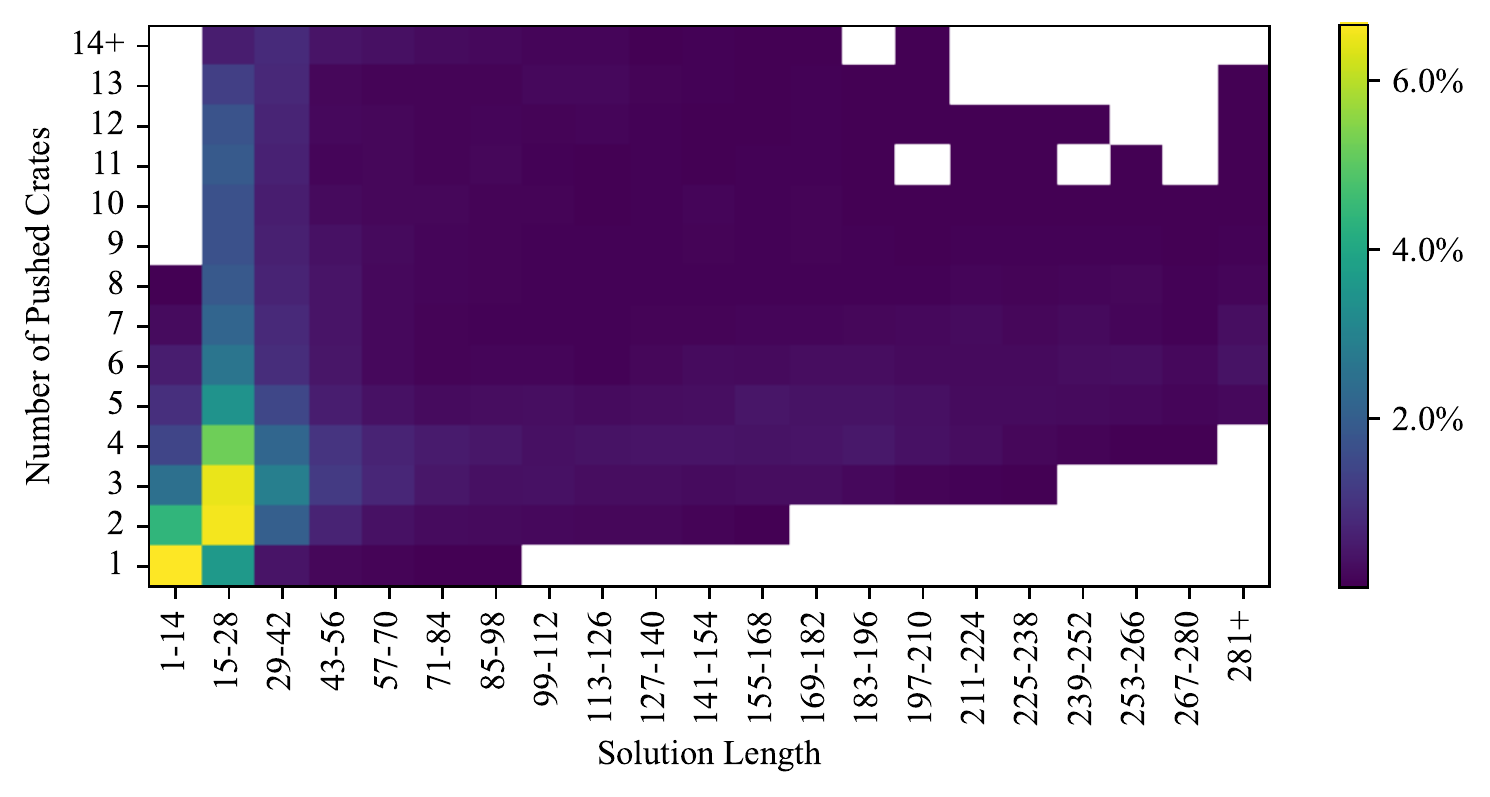}
		\caption{Multi-size GFlowNet (DS + PR + AUG)}
	\end{subfigure}
	
	\caption{The Expressive Ranges of the Sokoban Generators at the size $7 \times 7$. DS: Diversity Sampling. PR: Property Reward. AUG: Data Augmentation.}
	\label{fig:sokoban_er_comp}
\end{figure*}

Regarding playability for uncontrollable generation, C-PCGRL outperforms our generators given 1 trial, but given 10 trials, our generators significantly outperforms C-PCGRL while being $29.57\times$ faster. Regarding diversity, both methods are on-par in many aspects. Our generators' tile diversity is higher, while C-PCGRL's unique signature percentage is higher. In both cases, the difference is not statistically significant. \figurename\ \ref{fig:sokoban_er_comp} shows that our generators have broader expressive ranges. Along the pushed crates axis, C-PCGRL ranges expand up to $6.8\pm 0.45$ pushed crates, while our generators' ranges expands to $16.8\pm 1.92$ pushed crates ($p\text{-value}=0.97\%$). Along the solution length axis, C-PCGRL ranges expand up to $147.8\pm 33.9$ steps , while our generators' ranges expands to $332.0\pm 77.8$ steps ($p\text{-value}=0.79\%$). Regarding controls, our generators exhibits lower mean absolute error, higher $R^2$ and higher control scores compared to C-PCGRL for every control.

Regarding the training time, our generators are $10.06\times$ faster to train. And regarding the controllable generation time, our generators are $9.69\times$ faster (for 10 trials). There are multiple reasons for such a large difference in speed. As described in Section \ref{ssec:time}, we measured our model's call time, which turned out to be $0.38 wh + 0.20$ ms on average on the RTX 3090 GPU, so a single $7\times 7$ level needs $18.61$ ms. For C-PCGRL, a single call of the agent's model takes $0.54$ ms, and to generate a level, the episode length is $1917.24\pm 126.29$ steps (calculated from the generated sample), so the total generation time is $1035.31\pm 233.87$ ms ($55.63\times$ longer than our generators). Additionally, invoking the Sokoban solver is time consuming. Given 10 trials, our generators invoke the solver 10 times only. On the other hand, C-PCGRL agents require feedback from the solver during generation, and for generating the samples, they invoked the solver $16.02\pm 2.38$ times. Finally, our training process utilizes experience replay, which increases the sample efficiency.

C-PCGRL agents and our generators differ in some other aspects. C-PCGRL is an iterative generator, so it is compatible with mixed initiative co-creation \cite{Delarosa2021MixedInitiativeLD}. On the other hand, our generators are compatible with use cases where the level size can be controlled by the user without retraining.

\section{Conclusion}\label{sec:conc}

This paper presents a novel approach to train level generators without training data or shaped rewards by learning at multiple sizes. By utilizing the denser feedback at smaller sizes, our generators can learn without shaped rewards, which consume effort to design and require game-specific domain knowledge. We apply our approach to train an auto-regressive GFlowNet for controllable level generation. By using a recurrent architecture, our generator can fit different level sizes without changing its architecture, and can be globally consistent, which is required to generate diverse playable levels for Sokoban, Zelda, and Danger Dave. The results show that our generators create diverse playable levels based on user-supplied controls for various sizes (including out-of-training sizes) and a variety of games (Sokoban, Zelda, and Danger Dave). Compared to Controllable PCGRL for Sokoban level generation, our generators are $9\times$ faster at training and level generation, while also exhibiting lower control errors, and covering a significantly wider expressive range.




\bibliographystyle{elsarticle-num} 
\bibliography{references}


%\begin{thebibliography}{00}
%%% Text of bibliographic item
%
%\bibitem{}
%
%\end{thebibliography}
\end{document}